\documentclass[a4paper,fleqn]{cas-dc}

\usepackage[numbers]{natbib}
\usepackage{soul}
\soulregister\cite7
\sethlcolor{yellow}
\usepackage{colortbl}
\usepackage{xcolor}
\usepackage{amssymb}
\usepackage{makecell}

\usepackage{tcolorbox}
\tcbset{
  highlight/.style={
    colback=yellow,
    colframe=yellow,
    boxrule=0pt,
    left=0pt, right=0pt, top=0pt, bottom=0pt,
    boxsep=1pt,
    sharp corners
  }
}
\def\tsc#1{\csdef{#1}{\textsc{\lowercase{#1}}\xspace}}
\tsc{WGM}
\tsc{QE}
\begin{document}
\let\WriteBookmarks\relax
\def\floatpagepagefraction{1}
\def\textpagefraction{.001}
\let\printorcid\relax % 可去掉页面下方的ORCID(s)

\shorttitle{}    

\shortauthors{}  

\title [mode = title]{Embracing Intra-Class Heterogeneity for Semi-Supervised Medical Image Segmentation: From Diversity to Precision}  

\author[tju]{Yuqi Liu} 
\author[tju]{Yufei Chen}\cormark[1]
\author[tju]{Wei Fu}
\author[shu]{Xiaodong Yue} 
\author[cwu, cwu2]{Shuo Li}

\affiliation[tju]{organization={School of Computer Science and Technology, Tongji University},%Department and Organization
            % addressline={Address}, 
            city={Shanghai},
            postcode={201804}, 
            % state={Jiading},
            country={China}}
\cortext[1]{Corresponding author: yufeichen@tongji.edu.cn }  %声明通讯作者

\affiliation[shu]{
organization={Artificial Intelligence Institute, Shanghai University},
city={Shanghai},
postcode={200444},
country={China}
}

\affiliation[cwu]{
organization={Department of Computer and Data Sciences, Case Western Reserve University},
city={Cleveland},
postcode={44106},
state={OH},
country={USA}
}

\affiliation[cwu2]{
organization={Department of Biomedical Engineering, Case Western Reserve University},
city={Cleveland},
postcode={44106},
state={OH},
country={USA}
}

\begin{abstract}
Due to the scarcity of expert-annotated data, Semi-Supervised Medical Image Segmentation (SSMIS) has emerged as a promising approach. Many anatomical structures in medical images exhibit significant intra-class heterogeneity, with different regions showing heterogeneous intensity patterns within the same structure. However, existing methods inadequately exploit this intensity-manifested intra-class heterogeneity, resulting in uniform structural representations and imprecise segmentation. Furthermore, the scarcity of labeled data makes it more difficult to effectively capture such complex heterogeneity. 
To address this, we propose Multiple Prototype Contrastive Learning (MPCL), an SSMIS framework that possesses better diversity and better precision. It consists of three novel designs: 
First, we provide structural representations with better diversity and propose Intensity-aligned Heterogeneous Prototype Generation (IHPG) that effectively models intra-class heterogeneity by generating multiple prototypes aligned with intensity characteristics.
Second, we further enhance more diverse structural representations and build a solid foundation for more precise segmentation through Prototypical Space Optimization (PSO) that systematically optimizes a more discriminative and generalizable prototypical space.
Finally, we achieve segmentation results with better precision through Dual-branch Knowledge Alignment (DKA)  that efficiently promotes intra-class heterogeneity knowledge transfer from prototypical space to the segmentation network. 
Extensive experiments on three medical image datasets with significant intra-class heterogeneity  demonstrate that MPCL significantly outperforms existing methods, especially under extremely limited labeled data.
We believe this multiple prototype learning framework offers a novel paradigm for leveraging intra-class heterogeneity in medical images, which will provide an effective benchmark for SSMIS and advance the field of medical image analysis. Our code is publicly available at https://github.com/rhodaliu17/MPCL.
\end{abstract}

\begin{graphicalabstract}
\includegraphics[width=1\textwidth]{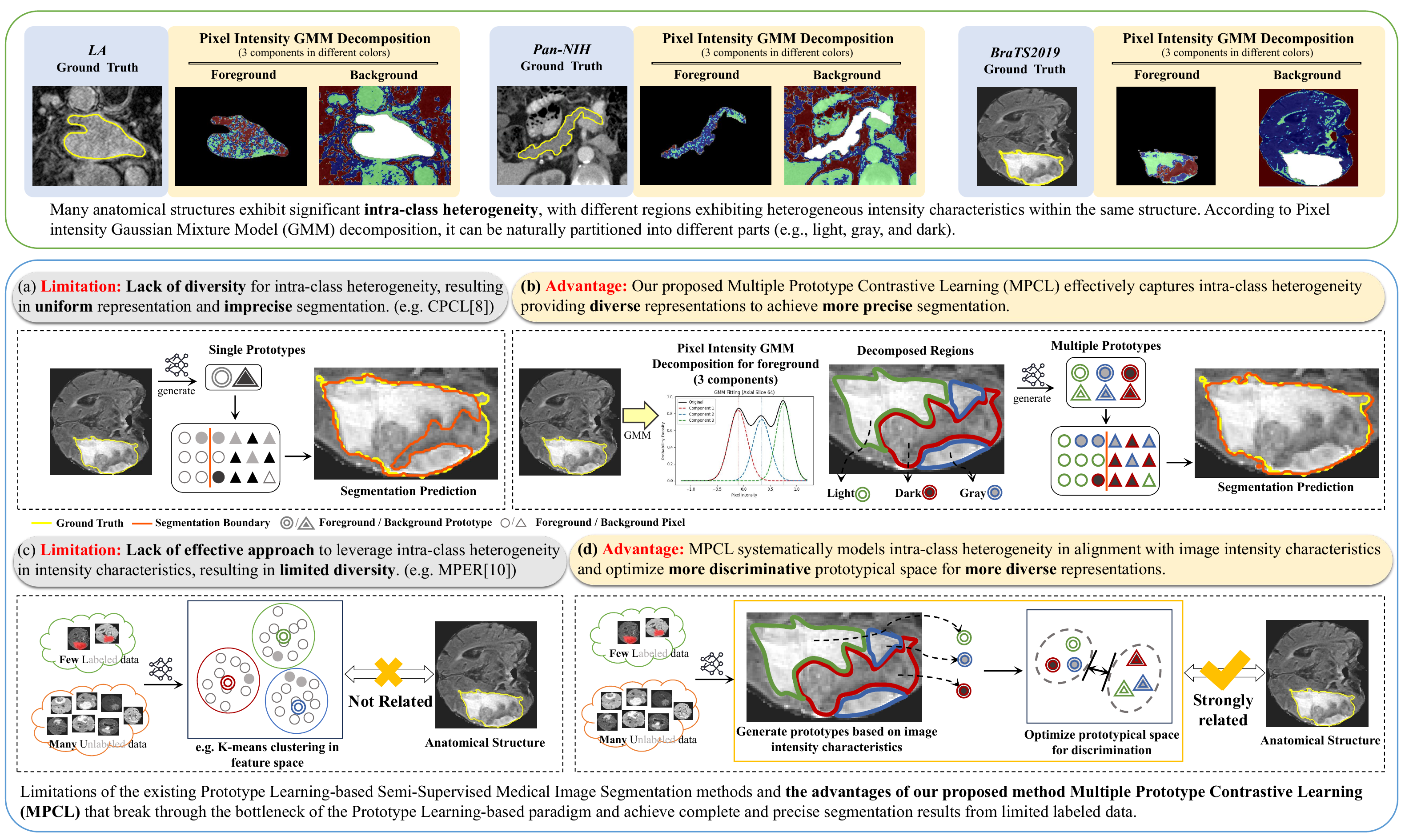}
\end{graphicalabstract}

\begin{highlights}
\item Propose a novel Multiple Prototypes paradigm, achieving more precise segmentation results.
\item Design IHPG to effectively model intra-class heterogeneity aligned with intensity characteristics.
\item Construct PSO to systematically optimize a more discriminative and generalizable prototypical space.
\item Apply DKA to efficiently transfer intra-class heterogeneity to the segmentation network.
\end{highlights}

%\nocite{*}

% Keywords

\maketitle

% Main text
\section{Introduction}
\begin{figure}
    \centering
    \includegraphics[width=1\linewidth]{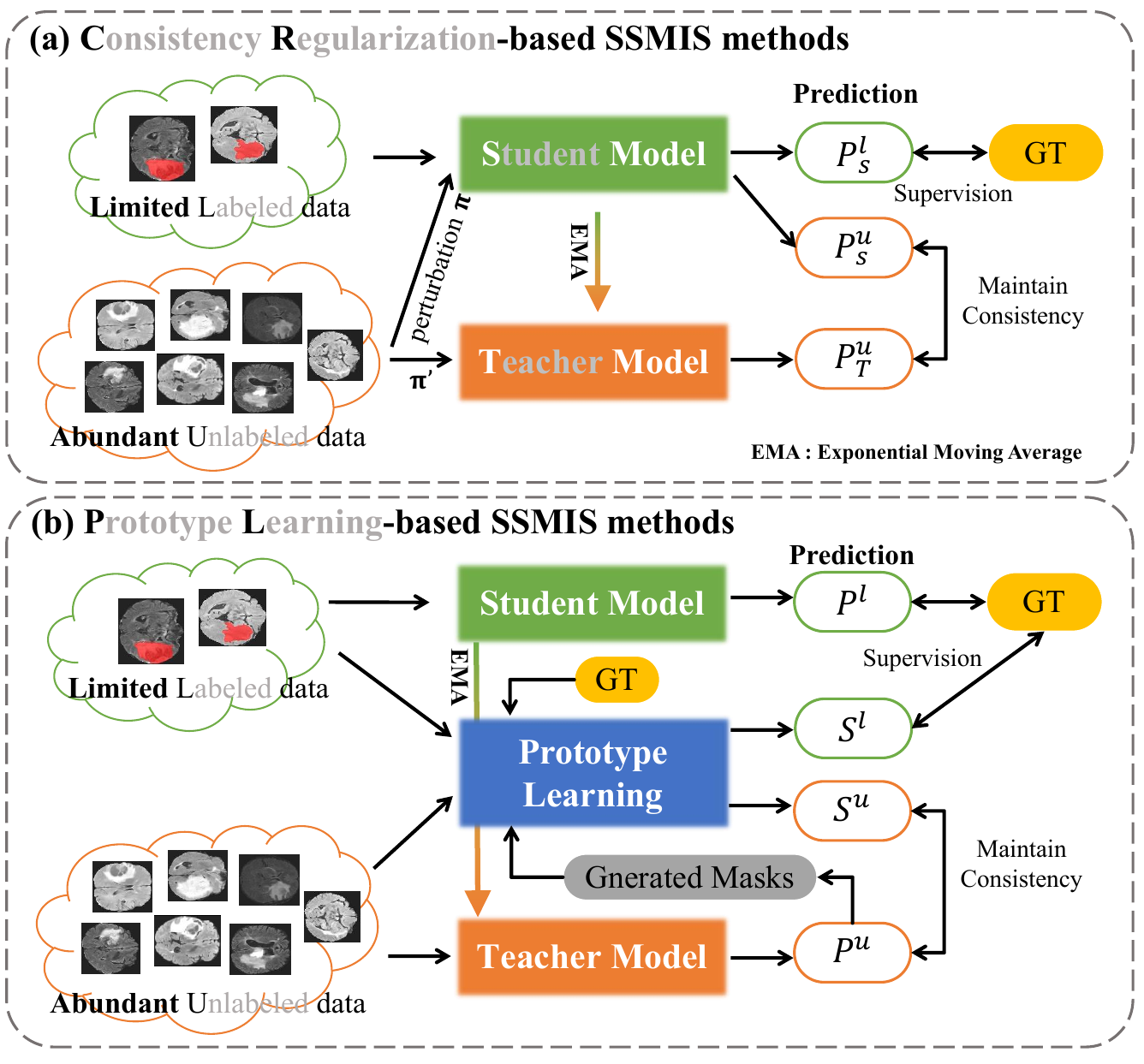}
    \caption{Paradigms of (a) Consistency Regularization-based (CR) SSMIS methods and (b) Prototype Learning-based (PL) SSMIS methods.}
    \label{firstfig}
    % \vspace{-15pt}
\end{figure}

Semi-Supervised Medical Image Segmentation (SSMIS) \cite{semireview2024WENG, 2024semireview, 2023semireview} has emerged to address the critical shortage of expert-annotated labels in medical image analysis and has developed rapidly with advances in deep learning. SSMIS leverages limited labeled data alongside abundant unlabeled data to perform segmentation across the entire dataset.
As shown in Fig. \ref{firstfig}(a), one of the most widely adopted paradigms is Consistency Regularization-based (CR) methods \cite{UAMT, DTC, TTMC}, which is based on the Mean-Teacher framework. In this paradigm, unlabeled data are subjected to different perturbations, and the model is trained to produce consistent segmentation predictions despite these perturbations.
Prototype learning-based (PL) methods \cite{upcol, CPCL, bapc, mper} have also gained increasing popularity in SSMIS, due to their ability to capture compact feature representations under limited labeled data—a capability well-demonstrated in few-shot learning. As shown in Fig. \ref{firstfig}(b), building upon the CR paradigm, PL paradigm introduces a novel consistency perspective compared to data-level perturbations. Consistency is maintained between predictions from the segmentation network and those derived from the prototype learning network. Notably, the quality of segmentation predictions improves proportionally with the discriminative power of the learned prototypes.

\begin{figure*}
    \centering
    \includegraphics[width=1\textwidth]{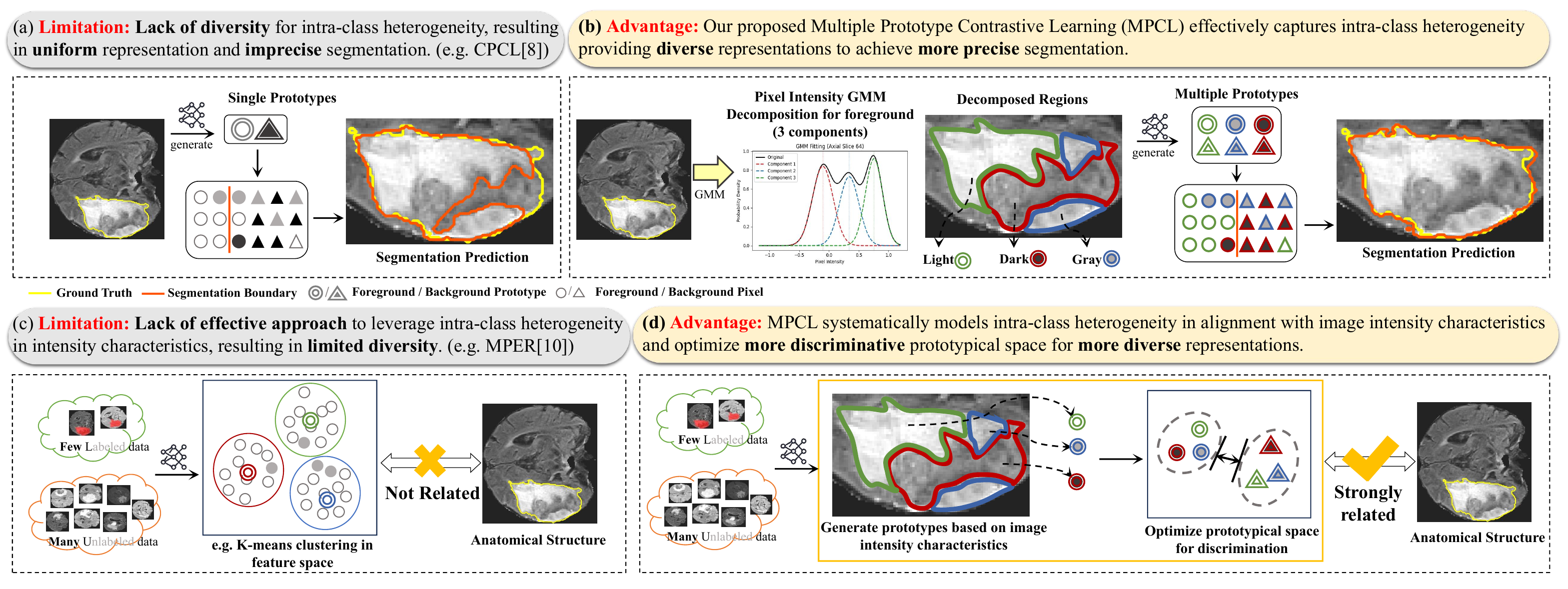}
    \caption{Limitations of the existing PL methods and the advantages of our MPCL that break through the bottleneck of the PL paradigm and achieve complete and precise segmentation results from limited labeled data. (a) and (c) are the limitations of the existing PL method. (b) and (d) are the corresponding solutions and advantages of our MPCL, respectively. (a) Existing methods only have single prototype for one representation. (b) Our MPCL has multiple prototypes to provide multiple representations and represent the intra-class heterogeneity. (c) Existing methods struggle to model heterogeneity. (d) Our MPCL systematically models intra-class heterogeneity according to image nature.}
    \label{limitationfigure}
\end{figure*}
However, there exists limitations [Fig. \ref{limitationfigure} (a), (c)] of the existing PL methods in SSMIS.

1) Lack of diversity for intra-class heterogeneity resulting in uniform representation and imprecise segmentation \cite{upcol,CPCL}.
Many anatomical structures exhibit significant intra-class heterogeneity, which is directly reflected in intensity. Specifically, different sub-regions within the same structure possess distinct underlying tissue properties, which are fundamentally manifested as differences in intensity characteristics\cite{intensity_lesion_ct, intensity_optimization}.
According to pixel intensity Gaussian Mixture Model (GMM) decomposition \cite{gmm, constrainedgmm, crossgmm} (Fig.~\ref{limitationfigure}(b)), these sub-regions can be naturally partitioned into different parts (e.g., light, gray, and dark).
However, existing SSMIS methods inadequately exploit intensity-manifested intra-class heterogeneity,
particularly within PL paradigms. Prototype learning can be regarded as voxel-level classification, where prototypes serve as standards for classifying each voxel into corresponding classes. However, most existing prototype-based methods learn a single representative prototype for each class, suffering from limited representation capability. By relying on only one prototype per class, these methods converge toward the dominant or uniform representation, failing to capture the intra-class heterogeneity. This limitation directly leads to incomplete segmentation in regions that differ from the learned prototype.

2) Lack of effective approach to leverage intra-class heterogeneity in intensity characteristics, resulting in limited diversity \cite{bapc,mper}.
Existing multiple prototype learning methods generate multiple prototypes relying mainly on feature-space generation without explicit relation with the intensity distribution within anatomical structures. In the feature space, voxels from high-density intensity regions contribute more samples, causing the generated prototypes to be inevitably dominated by these regions. By overlooking these intensity-based characteristics, minority intensity subregions are systematically underrepresented, failing to capture the full spectrum of diverse patterns within the same anatomical structure. Therefore, despite employing multiple prototypes, they ultimately lead to homogeneous representations that fail to reflect the true intra-class complexity of anatomical structures.

Motivation: To overcome the limitations of the existing PL paradigm and achieve diverse representations and precise segmentation with limited labeled data, we propose a novel PL framework for SSMIS: Multiple Prototype Contrastive Learning (MPCL). As illustrated in Fig.~\ref{limitationfigure}(b) and (d), MPCL not only effectively captures intra-class heterogeneity, providing diverse representations to achieve more precise segmentation, but also systematically models this heterogeneity in alignment with image intensity characteristics and optimizes more discriminative prototypical space for more diverse representations.
The framework comprises three novel designs: 

(1) Intensity-aligned Heterogeneous Prototype Generation (IHPG) effectively models intra-class heterogeneity by aligning with the heterogeneous intensity characteristics inherent in medical images, thereby providing more diverse structural representations. The intra-class heterogeneity present in medical images can be observed through intensity characteristics, where segmented regions can be partitioned into different parts (e.g. light, gray and dark) based on pixel intensities. Accordingly, we generate multiple prototypes for each segmented region to represent these characteristics while considering both semantic and spatial information. This generation originates from intensity characteristics and models the intra-class heterogeneity of medical images, providing representation with better diversity.

(2) Prototypical Space Optimization (PSO) systematically optimizes a more discriminative and generalizable prototypical space, so that we can further enhance more diverse structural representations and build a solid foundation for more precise segmentation. After generating multiple prototypes for both labeled and unlabeled data, it becomes essential to effectively leverage these diverse yet isolated representations. 
To this end, PSO optimizes the prototypical space through two mechanisms: 1) establishing clear decision boundaries between classes at the prototype level, and 2) integrating reliable supervision signals from labeled data with broad data distributions from unlabeled data. These two mechanisms synergistically refine the prototypical space, constructing more discriminative and generalizable representations that provide a solid foundation for more precise segmentation.

(3) Dual-branch Knowledge Alignment (DKA) efficiently transfers intra-class heterogeneity knowledge from the prototypical space to the segmentation network, leveraging intra-class heterogeneity knowledge for more precise segmentation. The prototypical space constructed by IHPG and PSO encodes rich intra-class heterogeneity knowledge and discriminative boundaries at the prototype level. However, there exists a gap between prototype-level representations and pixel-level segmentation predictions. Our DKA bridges this limitation through dual-branch knowledge alignment that explicitly aligns the prototype branch and the segmentation branch, enabling the intra-class heterogeneity knowledge to effectively guide pixel-level segmentation decisions, ensuring more precise segmentation results.

The main contributions of our method can be summarized as follows:
\begin{itemize}
\item We are the first to propose a multiple prototype-based learning framework for SSMIS that explicitly accounts for image intensity characteristics, providing a systematic mechanism to model the intra-class heterogeneity of anatomical structures in medical images.
\item We design a novel intra-class heterogeneity-aware generation strategy for generating multiple prototypes. By aligning with the inherent intensity characteristics, it models the full spectrum of intensity variations within each structure, thus effectively capturing intra-class heterogeneity and providing structural representations with better diversity.
\item We propose a novel prototype-level optimization learning strategy for optimizing prototypical space. It establishes more discriminative decision boundaries between classes while bridging complementary knowledge from labeled supervision and unlabeled data distributions, thus constructing a more discriminative and generalizable prototypical space that provides a solid foundation for precise segmentation.
\item We introduce a knowledge transfer strategy for transferring prototype knowledge to segmentation predictions. It ensures that the segmentation network inherits the intra-class heterogeneity knowledge and discriminative decision boundaries from learned prototypes, thus achieving more precise segmentation on anatomical structures.

\end{itemize}

In conclusion, our MPCL has three key advantages: 
a) Higher Precision: Compared with existing SSMIS methods that rely on single prototypes, our MPCL achieves more precise segmentation results by effectively embracing intra-class heterogeneity through multiple prototype learning. 
b) Fewer Labels: While maintaining higher precision, our MPCL demonstrates superior performance even under limited supervision. By capturing diverse intra-class patterns through multiple prototypes, our method makes  full  use of labeled and unlabeled data and achieves competitive results.
c) Great Applicability: Unlike methods designed for specific organs or modalities, our MPCL leverages the universal intensity characteristics inherent in medical images, making it broadly applicable across diverse anatomical structures and imaging protocols.

\section{Related Works}
\subsection{Semi-Supervised Medical Image Segmentation}
Medical image segmentation \cite{rednet,chen2022target} plays a crucial role in medical image analysis \cite{segreview2024,dedl}, serving multiple clinical applications including diagnosis, treatment planning, and disease monitoring. However, the scarcity of labeled data poses a significant challenge. Under such circumstances, Semi-Supervised Medical Image Segmentation (SSMIS) \cite{semiMIA1, semiMIA2} has emerged as an effective solution to leverage abundant unlabeled data alongside limited annotations. Recently, Consistency Regularization-based (CR) methods have gained prominence in SSMIS \cite{UAMT, urpc, fussnet}. It originates with the $\Pi$-model \cite{gaussianwarmup}, which applies different perturbations to input data while enforcing prediction consistency. This concept is further advanced by Temporal Ensembling \cite{gaussianwarmup}, which utilizes an Exponential Moving Average (EMA) of predictions as consistency targets. Mean Teacher (MT) \cite{meanteacher} subsequently revolutionizes this direction by introducing an architecture comprising student and teacher networks. Unlabeled data with various perturbations are processed through both networks, with consistency maintained between their respective predictions. Expanding on this, there have arisen numerous innovative consistency strategies \cite{DTC2, DTC3, DTC4}. For instance, Luo et al. \cite{DTC} employ dual prediction targets, including segmentation predictions and signed distance field map predictions, and ensure their consistency through bidirectional transformations. Building on this, Chen et al. \cite{TTMC} incorporate another prediction target, signed attention map predictions, and enforce consistency across all three prediction types. Similarly, Wu et al. \cite{mcnet} utilize two decoders to generate distinct segmentation outputs while encouraging consistency between them.
These CR methods have demonstrated significant progress in SSMIS by effectively exploiting unlabeled data through various consistency constraints.

\subsubsection{Single Prototype Learning}
Prototypes initially gained widespread utilization in few-shot learning due to their efficient capability to represent data features. Dong et al. \cite{profs2} first apply prototypes in few-shot segmentation.  Furthermore, Li et al. \cite{psanet} implement a salient attention strategy to enhance prototype feature representation. Given these advances, prototype learning has subsequently been incorporated into semi-supervised segmentation frameworks \cite{upcol, CPCL, LIUpro}. Xu et al. \cite{nipspro} explore consistency relationships between prototype-based predictions and segmentation outputs. Additionally, Xu et al. \cite{CPCL} perform cyclic prototype consistency learning, which leverages prototypes from labeled data to generate predictions on unlabeled data, while implementing reverse operations for unlabeled data. 
They further maintain the consistency between the predictions generated from the prototype and the segmentation results. 
Lu et al. \cite{upcol} further innovate by integrating information from both labeled and unlabeled sources while enforcing consistency between prototype-based and segmentor predictions. 
These developments demonstrate that prototype learning provides an efficient and intuitive way for feature representation. However, we find it insufficient to represent the complex internal diversity in anatomical structures using only a single prototype. Therefore, we introduce multiple prototypes to effectively capture the intra-class heterogeneity for more diverse representations and further enhance the segmentation precision.

\subsubsection{Multiple Prototype Learning}
Addressing the inherent limitations of single prototype representation, recent works have explored Multiple Prototype Learning (MPL) methods \cite{bapc, mper, ASGNET, implicit, mvpcl}. Tian et al. \cite{implicit} generate a set of prototypes for each image in a mini-batch, and the final prediction results are obtained by computing and voting across multiple sets of prototypes. Li et al. \cite{mvpcl} generate one prototype per view under multiple view scenarios to form multiple prototypes and achieve enhanced representation capabilities. The above multiple prototype methods still follow the main idea of single prototype learning, lacking diverse representations. Wang et al. \cite{bapc} focus on boundary variations and generate boundary-aware prototypes, including prototypes for boundary areas and for inner areas. However, this spatial-based feature decomposition assumes homogeneity within each decomposed region, resulting in under-segmentation of complex heterogeneous subregions. Bi et al. \cite{mper} explore intra-class variations by clustering voxels into multiple prototypes through MiniBatchKMeans in the feature space. However, the prototypes generated by generic feature-space clustering without intensity guidance inevitably gravitate toward dominant patterns, failing to segment heterogeneous subregions.
In conclusion, the potential of multiple prototypes to effectively capture intensity-related intra-class heterogeneity is inadequately exploited in MPL SSMIS methods, leaving representation diversity fundamentally limited.

\subsubsection{Discussion for Related Works}
The fundamental objective of SSMIS is to achieve precise segmentation with minimal labeled supervision. Within the PL paradigm in SSMIS, prototype diversity is crucial for capturing the complete characteristics of anatomical structures, which directly determines segmentation precision. Specifically, diverse prototypes enable the model to capture the multifaceted characteristics of anatomical structures, including regional variations in tissue composition, morphological heterogeneity, and boundary transitions. This comprehensive representation is essential for accurate tissue delineation, as it allows the segmentation network to distinguish subtle intra-class differences. Although some existing works have attempted to employ multiple prototypes to capture different representations, none of them  has systematically exploited the intrinsic intensity characteristics inherent in medical images to model intra-class heterogeneity, particularly within the PL paradigm. To address this, we introduce MPCL, which leverages IHPG, PSO, and DKA to collectively generate multiple prototypes with more diverse representations, achieving more precise segmentation results in SSMIS.

\section{Methods}
The proposed MPCL framework (Fig. \ref{architecture}) achieves precise Semi-Supervised Medical Image Segmentation (SSMIS) (Sec. \ref{sec:pre}) by embracing intra-class heterogeneity through multiple prototypes. Specifically, it consists of three designs: (1) Intensity-aligned Heterogeneous Prototype Generation (IHPG)  effectively models intra-class heterogeneity aligned with intensity characteristics, providing representations with better diversity. (see Sec. ~\ref{sec:ihpg}). (2) Prototypical Space Optimization (PSO) systematically optimizes prototypical space to further enhance diversity, establishing a solid foundation for more precise segmentation. (see Sec. ~\ref{sec:dpo}). (3) Dual-branch Knowledge Alignment (DKA) efficiently transfers intra-class heterogeneity knowledge, achieving segmentation results with better precision. (see Sec. ~\ref{sec:DKA}).

\subsection{Preliminaries}
\label{sec:pre}
\begin{figure*}
    \centering
\includegraphics[width=1\linewidth]{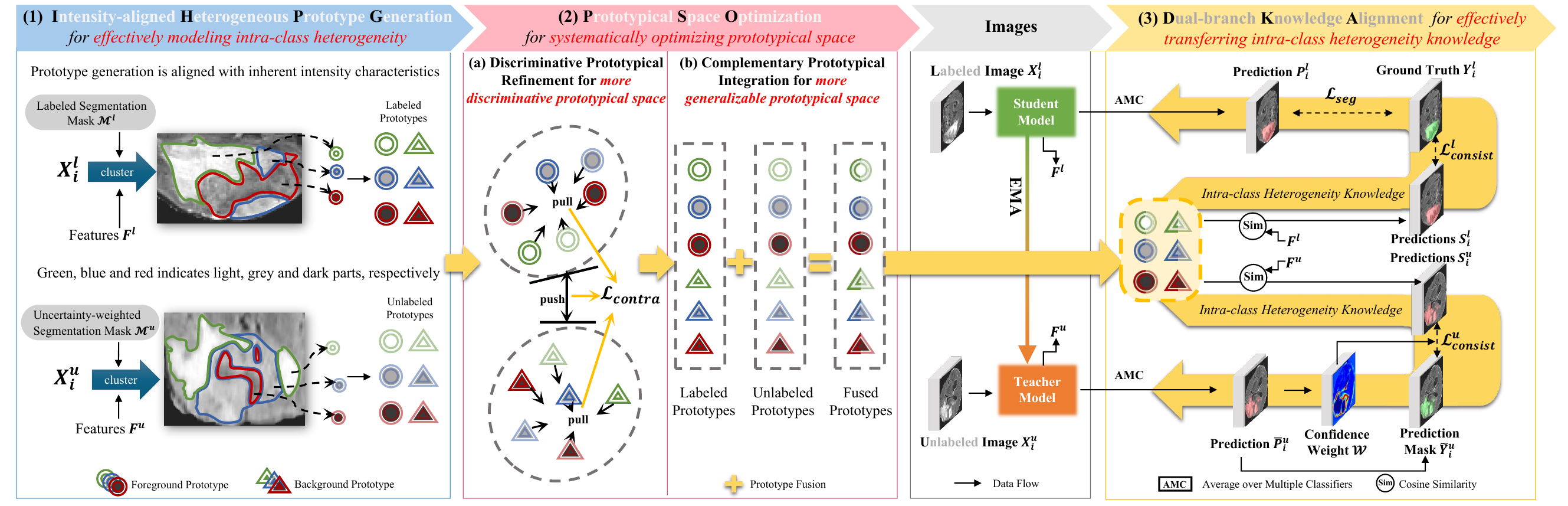}
    \caption{Overflow of our proposed method Multiple Prototypes Contrastive Learning (MPCL), which consists of three designs, including Intensity-aligned Heterogeneous Prototype Generation (IHPG) providing representations with better diversity via effectively modeling intra-class heterogeneity guided by intensity characteristics, Prototypical Space Optimization (PSO) systematically establishing a solid foundation for more precise segmentation via optimizing prototypical space to further enhance diversity and Dual-branch Knowledge Alignment (DKA) achieving segmentation results with better precision via efficiently transferring intra-class heterogeneity knowledge.}
    \label{architecture}
\end{figure*}

SSMIS is a challenging task that aims to achieve accurate medical image segmentation across the entire dataset using only limited labeled data and abundant unlabeled data during training. 
We adopt the commonly used Mean-Teacher architecture following \cite{upcol, CPCL} (Figs. \ref{firstfig} and \ref{architecture}). 
The student model is trained through loss functions while the teacher model’s parameters are updated by the Exponential Moving Average (EMA) of the student model’s parameters. For each model, we use the Average over Multiple Classifiers (AMC) strategy as in \cite{upcol} to enhance its robustness, 
which contains four segmentation heads each supervised by a distinct loss function (CE, Focal, Dice, and IoU, respectively) to independently predict complementary results.

The training dataset is defined as $\boldsymbol{\mathcal{D}} = \{\mathcal{D}^{l}, \mathcal{D}^{u}\}$, where $\mathcal{D}^{l}=\{X^l_i, Y^l_i\}^{N}_{i=1}$, $\mathcal{D}^{u} =\{X_{i}^{u}\}_{i=N+1}^{N+M}$, and $N$ is the number of labeled data, $M$ is the number of unlabeled data, with $N \ll M$. 
We define the prediction mask $\tilde{Y}_{i}^{u}$ as:
\begin{equation}
    \tilde{Y}_{i}^{u}=\underset{c \in\{0, 1 ...,C-1\}}{\arg \max }\bar{P}_{i,c}^{u}, \quad \bar{P}_{c}^{u}=\frac{1}{G} \sum_{g=1}^{G} P_{g,c}^{u},
\end{equation}
where $P_{g,c}^{u}$ is the prediction of class $c$ in the $g$-th segmentation head. Since $\tilde{Y}_{i}^{u}$ inevitably contains uncertain predictions, the uncertainty at each voxel position $x$ is estimated via entropy as:
\begin{equation}
    unc_x=-\sum_{c=0}^{C-1} \bar{P}_{c}^{u} \log \bar{P}_{c}^{u}.
\end{equation}
It can be observed that high uncertainty values correspond to less reliable predictions, while low uncertainty values indicate more confident predictions. To alleviate the effect of unreliable predictions, we design a confidence weight $\mathcal{W}_x$ at each voxel position $x$ for subsequent processes as:
\begin{equation}
\label{weight}
\mathcal{W}_{x} = 1-\frac{unc_{x}}{\sum_{x=1}^{H \times W \times D} unc_{x}}.
\end{equation}

\subsection{Intensity-aligned Heterogeneous Prototype Generation (IHPG)}
\label{sec:ihpg}

Our IHPG effectively models intra-class heterogeneity by generating multiple prototypes aligned with intensity characteristics, providing more diverse representations of anatomical structures. It comprises two stages:

1) \textbf{Intensity-Driven Initialization}:
It ensures prototypes are initialized aligned with inherent intensity characteristics, enabling efficient modeling of intra-class heterogeneity from the beginning. It provides diverse initialization of the anatomical structures for subsequent refinement.

For each class in the input medical image, we aim to generate $K$ representative prototypes $\mathcal{P}$
that capture intra-class heterogeneity within the anatomical structure.
Let $X$ denote the input image and $\mathcal{M}$ denote the segmentation mask. $X$ is normalized with intensity values scaled to $[0, 1]$ per image as $\mathbf{X}$, preserving intra-image relative intensity distributions while  eliminating inter-image intensity variations. To capture the intensity heterogeneity within each class, we construct a histogram of the image intensities within the corresponding masked region. Specifically, we compute the histogram over $B$ bins as follows:
\begin{equation}
\text{hist}(b) = \sum_{x  \in \mathcal{M} }
\mathbb{1}\left[(\mathbf{X}_i(x)\cdot B) = b\right], 
\quad b \in \{0, 1, \ldots, B-1\},
\end{equation}
where $\mathbb{1}[\cdot]$ is the indicator function, and $\mathbf{X}_i(x)$ denotes the normalized intensity at each voxel position $x$. We then identify local maxima in the histogram as intensity peaks, which represent distinct intensity patterns within the structure:
\begin{equation}
\mathcal{P}_{\text{peaks}} = \{b \mid \text{hist}(b) > \text{hist}(b-1) 
\wedge \text{hist}(b) > \text{hist}(b+1)\}.
\end{equation}

To ensure diversity among selected peaks, we apply non-maximum suppression (NMS) to filter peaks that are too close in intensity space. The peaks are then sorted by their histogram counts in descending order, and the top $K$ peaks are selected.
These selected peaks provide initialization positions that naturally align with intensity-driven heterogeneity in medical images. For each selected peak, 
we randomly choose one voxel from the corresponding intensity bin to form the initial intensity-guided prototype $\mathcal{P}^0$.

2) \textbf{Feature-Spatial Clustering}:
It refines the initialized prototypes to achieve comprehensive modeling of intra-class heterogeneity by integrating both semantic and spatial information. 
It constructs more complete representations grounded in the intrinsic feature distribution of the image.

To begin with, we incorporate positional information by concatenating the coordinates of all voxels with the corresponding feature map $\mathbf{F}$. 
Subsequently, we apply the segmentation mask $\mathcal{M}$ to extract valid features $\mathbf{F'}$.
We then progressively refine the initial intensity-guided prototype $\mathcal{P}^0$ through an iterative procedure, yielding the final prototypes $\mathcal{P}\in \mathbb{R} ^{Ch\times K}$, where $Ch$ denotes the channel number. Here, we define a modified distance function $Dist$ that considers both feature and spatial information, as follows:
\begin{equation}
\label{eqptdis}
    % D=\sqrt{(d_f)^2+(d_v/r)^2}
    Dist=\sqrt{d_f+d_v/r},
\end{equation}
where $d_f$ and $d_v$ represent the Euclidean distances for feature maps and voxel coordinates, with $r$ balancing their contributions. 
During each iteration $t$, we compute an association map $Q$ that quantifies the relationship between $x$-th voxel feature $f'_{x}$ in valid features $F'$ and $k$-th prototype $\mathcal{P}_k$ according to our defined distance function $Dist(\cdot)$, as follows:
\begin{equation}
    Q^{t}_{x,k}=e^{-Dist(f'_{x},\mathcal{P}^{t-1}_{k})}.
\end{equation}

Based on this, we proceed to update the prototypes through the weighted sum of masked features as follows:
\begin{equation}
\label{eqptupdate}
    \mathcal{P}_k^t = \frac{1}{\sum_{x=1}^{N_m}Q^{t}_{x,k}} \sum_{x=1}^{N_m} Q_{x,k}^t f'_{x}.
\end{equation}

For labeled data, the student model generates $K$ labeled prototypes $\mathcal{P}^l_c=\{\mathcal{P}^l_{c,1}, ..., \mathcal{P}^l_{c,K}\}$ per class $c$ using the 3rd-layer decoder features $\mathbf{F}^l$ and ground truth mask $\mathcal{M}^l = Y^l$. For unlabeled data, the teacher model generates $K$ unlabeled prototypes $\mathcal{P}^u_c=\{\mathcal{P}^u_{c,1}, ..., \mathcal{P}^u_{c,K}\}$ per class $c$ using features $\mathbf{F}^u$ and uncertainty-weighted segmentation mask $\mathcal{M}^u$, defined as:
    \begin{equation}
    \mathcal{M}^u_{i,x} = \mathcal{W}_{x} \tilde{Y}_{i,x}^{u},
\end{equation}
where $\mathcal{W}_x$ from Eq.~(\ref{weight}) suppresses unreliable voxel predictions by weighting each position according to its confidence, ensuring that prototype generation is guided by more reliable regions of the unlabeled predictions.

\textbf{Discussion of the Innovation}: Our IHPG proposes a novel intra-class heterogeneity-aware generation strategy for multiple prototypes, which aligns with the inherent intensity characteristics and models the full spectrum of intensity variations within each structure. Therefore, it effectively models intra-class heterogeneity of the structure, capturing structural representations with better diversity.

\subsection{Prototypical Space Optimization (PSO)}
\label{sec:dpo}
After IHPG, our PSO systematically optimizes the generated prototypical space to be more discriminative and generalizable, which further enhances more diverse structural representations and builds a solid foundation for more precise segmentation.  It consists of two mechanisms:

1) \textbf{Discriminative Prototypical Refinement}: It establishes clear decision boundaries between classes at the prototype level in the generated prototypical space, refining a more discriminative prototypical space. It optimizes inter-class discriminability while maintaining intra-class heterogeneity.

Let $\mathcal{P} = \{\mathcal{P}^{l}\} \cup \{\mathcal{P}^{u}\}$ 
denotes the complete prototype set including both labeled and unlabeled 
prototypes, and $\mathcal{P}_c$ denotes the prototype set of class $c$.
The representation optimization in the prototypical feature space is achieved via contrastive loss as:
\begin{equation}
    \mathcal{L}_{contrast} = -\frac{1}{2K\cdot C} \sum_{c=1}^{C} 
    \sum_{k=1}^{2K} \log \frac{PT^+_{c,k}}{PT^+_{c,k} + PT^-_{c,k}}, 
\end{equation}
in which $PT^+_{c,k}$ and $PT^-_{c,k}$ are defined as:
\begin{equation}
\begin{aligned}
    PT^+_{c,k} &= \sum_{\mathcal{P}_j \in \mathcal{P}_c \setminus \{\mathcal{P}_{c,k}\}} \exp(\text{cos}(\mathcal{P}_{c,k}, \mathcal{P}_{j})/\tau), \\
    PT^-_{c,k} &= \sum_{\mathcal{P}_j \in \mathcal{P} \setminus \mathcal{P}_c} \exp(\text{cos}(\mathcal{P}_{c,k}, \mathcal{P}_{j})/\tau),
\end{aligned}
\end{equation}
where $\text{cos}(\cdot)$ is the cosine similarity, and $\tau$ is the 
temperature parameter that controls the sharpness of the distribution 
and is set to $0.07$ as commonly recommended.

2) \textbf{Complementary Prototypical Integration}: It integrates reliable supervision from labeled data with broader distributional knowledge from unlabeled data to bridge the gap between data sources. It further enriches the prototypical space to learn more generalizable representations.

For $k$-th prototype of one class $c$, the fused prototype $\mathcal{P}^{f}_{c,k}$ is obtained by:
\begin{equation}
    \mathcal{P}^{f}_{c,k} = (\mathcal{P}^{l}_{c,k} + \lambda \mathcal{P}^{u}_{c,k})/(1+\lambda),
\end{equation}
where $\mathcal{P}^l_{c,k},\mathcal{P}^u_{c,k}$ respectively represent the $k$-th labeled and unlabeled prototypes of class $c$. As training advances, unlabeled prototypes generated by the teacher model progress from initially unreliable to increasingly trustworthy representations. 
Therefore, we adopt the widely-used time-dependent Gaussian warming-up function $\lambda$ to dynamically control its contribution in the fusion process \cite{CPCL, gaussianwarmup},  defined as:
\begin{equation}
\label{gussiancompute}
    \lambda(s)=e^{(-5(1-s /s_{max})^{2})},
\end{equation}
where $s$ is the current training step and $s_{max}$ is the maximum training step. It remains small in early training to suppress unreliable contributions and increases rapidly as the network stabilizes.
The fusion process is applied to every corresponding labeled and unlabeled prototype across each class, generating the final fused prototypes $\mathcal{P}^{f}_c = \{\mathcal{P}^{f}_{c,1}, \mathcal{P}^{f}_{c,2}, ..., \mathcal{P}^{f}_{c,K}\}$.
% (functionality)
This fusion is a collaborative knowledge representation strategy that can capture both explicit supervisory information in limited labeled data and implicit structural information in large amounts of unlabeled data. Labeled prototypes provide reliable guidance in expert annotations, while unlabeled prototypes offer a broader perspective on the underlying data distribution. These complementary information sources enhance each other to form a more robust and generalizable prototype feature representation, constructing the final prototypical feature space for subsequent segmentation.

\textbf{Discussion of the Innovation}: Our PSO proposes a novel prototype-level optimization learning strategy, which not only establishes more discriminative decision boundaries between classes but also bridges the complementary knowledge from both reliable labeled supervision and broad unlabeled data distributions. It further enhances the diverse representations at the prototype level and builds a solid foundation for precise segmentation.

\subsection{Dual-branch Knowledge Alignment (DKA)}
\label{sec:DKA}

Our DKA efficiently transfers intra-class heterogeneity knowledge from the prototypical space to the segmentation network through bidirectional alignment between the prototype learning branch and the segmentation learning branch, thereby achieving more precise segmentation results.

To establish this alignment, we first obtain the predictions from the prototype branch using non-parametric metric learning without introducing additional learnable parameters.
Specifically, we compute the cosine similarity between $x$-th voxel feature $f_x$ in the feature map $\mathbf{F}$ and each prototype $\mathcal{P}^{f}_{c,k}$. A higher similarity value indicates a stronger association between the voxel and the corresponding class of the prototype. Based on these similarity computations, we formulate voxel-wise prediction maps for both labeled and unlabeled data, as follows:
\begin{equation}
S_{x,c}^{\beta} = \max_{k \in {1,...,K}} \left( \frac{f_{x}^{\beta} \cdot \mathcal{P}^{f}_{c,k}}{\max \left(\left\|f_{x}^{\beta}\right\|_{2} \cdot\left\|\mathcal{P}^{f}_{c,k}\right\|_{2}, \epsilon\right)}\right),
\end{equation}
where $\epsilon$ is fixed to $1e^{-8}$; $\beta \in\{l, u\}$ denotes labeled and unlabeled data, with $S^l$ and $S^u$ as their corresponding prototype-based predictions, respectively.

These prototype-based predictions capture rich intra-class heterogeneity information derived from our multiple prototype representations. Based on this, dual-branch consistency constraints are enforced between these predictions and the segmentation network outputs, explicitly aligning the two branches in a bidirectional manner. This alignment not only transfers intra-class heterogeneity knowledge to guide pixel-level segmentation decisions, but also allows segmentation outputs to regularize the prototype learning branch against drift from segmentation-relevant semantics. This mutual refinement collectively strengthens the framework's capacity to characterize complex anatomical structures, ultimately achieving more precise segmentation results.
% thereby enhancing the network's capability to characterize complex anatomical structures.
Accordingly, the consistency constraint for labeled data is enforced via labeled consistency loss:
\begin{equation}
\mathcal{L}_{consist}^{l}= \mathcal{L}_{CE}(S_{}^{l},Y_{}^{l}).
\end{equation}
For unlabeled data, the consistency constraint is enforced via unlabeled consistency loss:
\begin{equation}
\mathcal{L}_{consist}^{u}=\frac{1}{H\times W\times D} \sum_{x=1}^{H\times W\times D}\mathcal{W}_{x}\cdot\mathcal{L}_{CE}(S_{x}^{u},\tilde{Y}^{u}_{x}),
\end{equation}
where $\mathcal{W}_x$ is defined in Eq.~(\ref{weight})  to reduce the influence of unreliable predictions.

Additionally, to optimize the segmentation network toward accurate predictions, we define the supervision segmentation loss $\mathcal{L}_{seg}$ on labeled data. Leveraging the AMC strategy, it comprises the equally weighted average of four loss functions, including cross-entropy loss (CE), focal loss \cite{focalloss}, dice loss \cite{vnetdiceloss}, and IoU loss \cite{iouloss}, defined as:
\begin{equation}
\begin{split}
\mathcal{L}_{seg} = \frac{1}{4}(&\mathcal{L}_{CE}(P^{l}_{1}, Y^{l}_{})+\mathcal{L}_{FOCAL}(P^{l}_{2}, Y^{l}_{}) \\
&+\mathcal{L}_{DICE}(P^{l}_{3}, Y^{l}_{})+\mathcal{L}_{IOU}(P^{l}_{4}, Y^{l}_{})).
\end{split}
\end{equation}

The overall optimization objective of our framework is therefore formulated as:
\begin{equation}
\mathcal{L} = \mathcal{L}_{seg} + \mathcal{L}_{consist}^{l}+\lambda(\mathcal{L}_{consist}^{u}+ \mathcal{L}_{contrast}),
\end{equation}
where $\lambda$ is the time-dependent Gaussian warming-up function defined in Eq.~(\ref{gussiancompute}). In early training, $\mathcal{L}_{seg}$ and $\mathcal{L}^l_{consist}$ provide reliable supervision signals to guide the training process. Since both $\mathcal{L}^u_{consist}$ and $\mathcal{L}_{contrast}$ depend on the reliability of unlabeled representations from the teacher model, applying the same $\lambda$ naturally suppresses their contributions in early training to accommodate network instability, and gradually activates them as the network stabilizes.

\textbf{Discussion of the Innovation}: 
Our DKA establishes a knowledge transfer strategy to bridge the prototypical and segmentation spaces. It explicitly aligns the prototype branch and the segmentation branch, transferring optimized intra-class heterogeneity knowledge encoded in prototype-based predictions to guide pixel-level segmentation decisions. Therefore, it enables more precise segmentation of anatomical structures.

\section{Experiments}
\subsection{Datasets}
\begin{figure}
    \centering
    \includegraphics[width=1\linewidth]{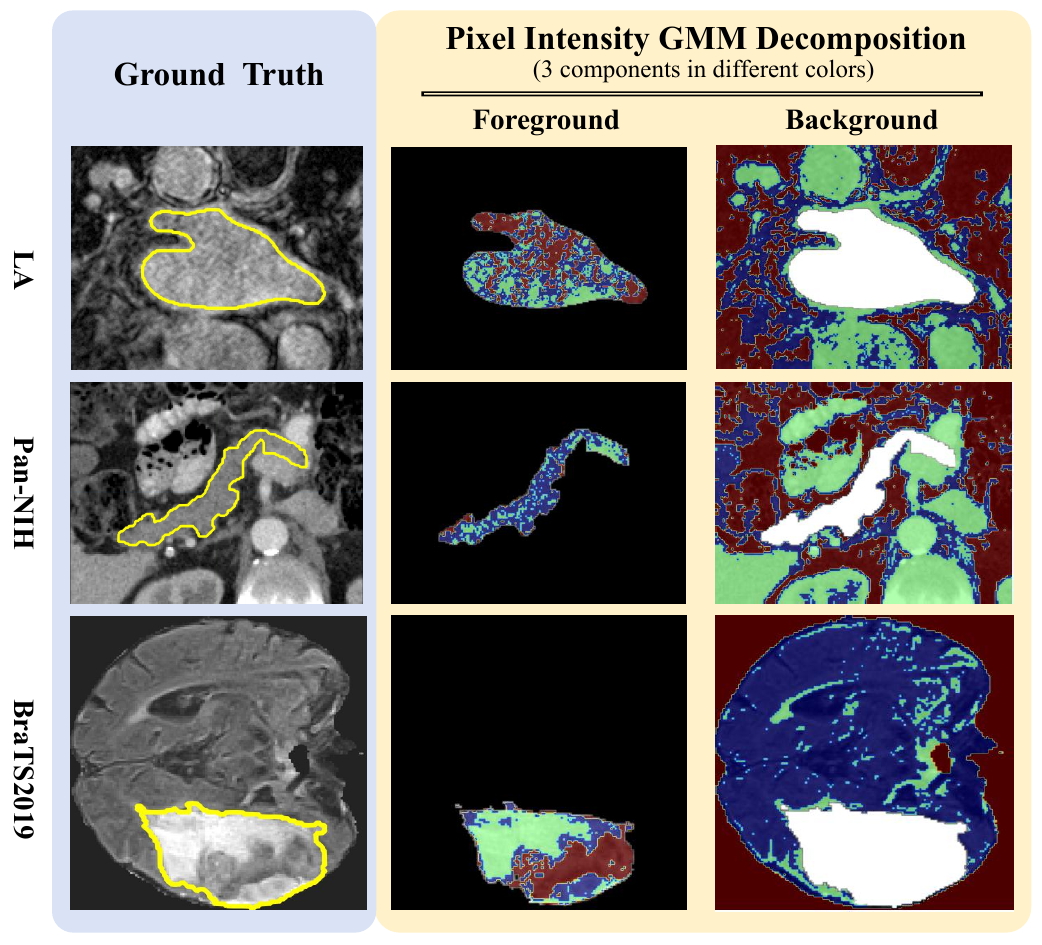}
    \caption{To demonstrate the improvement on segmentation precision via embracing intra-class heterogeneity, experiments are conducted on three datasets (LA, Pan-NIH, and BraTS2019) with intra-class heterogeneity on both foreground and background.}
    \label{gmmfigure}
\end{figure}

To demonstrate the effectiveness of the proposed method, our experiments were conducted on three different medical image datasets with intra-class heterogeneity, including Left Atrium (LA), NIH Pancreas CT (Pan-NIH) and BraTS2019 dataset, as shown in Fig. \ref{gmmfigure}. We apply Gaussian Mixture Models (GMM) to perform pixel intensity decomposition to intuitively exhibit heterogeneous intensity patterns within both foreground and background, where we empirically decompose them into $3$ distinct components with $3$ colors.

\textbf{The Left Atrium (LA) dataset} \cite{LA} contains 100 3D gadolinium-enhanced MR images from the 2018 Atrial Segmentation Challenge, with a consistent resolution of $0.625 \times 0.625 \times 0.625 mm^{3}$. \textbf{The NIH Pancreas CT (Pan-NIH) dataset} \cite{PAN} comprises 80 abdominal contrast-enhanced 3D CT scans from the National Institutes of Health Clinical Center, with resolutions of $512\times 512$ pixels and slice thickness ranging from $1.5$ to $2.5 mm$. \textbf{The BraTS2019 dataset} \cite{bratsdataset} includes 335 multimodal MRI scans of glioblastoma from the BraTS 2019 challenge, containing four modalities: native (T1), post-contrast T1-weighted (T1Gd), T2-weighted (T2), and T2 Fluid Attenuated Inversion Recovery (T2-FLAIR) volumes. We utilized T2-FLAIR for whole brain tumor segmentation due to its optimal visualization of the entire tumor region.

\subsection{Implementation Details and Metrics}
1) Implementation Details: We conducted all experiments on a single NVIDIA GeForce RTX 3090 GPU. We utilized VNet \cite{vnetdiceloss} as the backbone and trained for 10k iterations using Adam optimizer with a learning rate 0.001. The batch size was set to 2, including 1 labeled image and 1 unlabeled image. For data augmentation, we applied random cropping, flipping, and rotation following previous works \cite{TTMC,upcol, CPCL}. 
The hyperparameter $r$ in Eq. \eqref{eqptdis} was set to 100 following the implementation of \cite{ASGNET}.
% The hyperparameter $r$ in Eq. \eqref{eqptdis} was set to 100 following \cite{ASGNET}.
\textbf{For LA and Pan-NIH dataset}, we used the same data partition as in \cite{TTMC,upcol}, dividing the LA datasets into 80 training and 20 validation images, and the Pan-NIH dataset into 60 training and 20 validation images. For LA dataset, we center-cropped all images at the heart region and normalized them to zero mean and unit variance. In training, images were randomly cropped into patches of size $112 \times 112 \times 80$, and the same size with a stride of $18 \times 18 \times 4$ in inference. For Pan-NIH dataset, the images were cropped centering on the pancreas with enlarged margins (25 voxels) and normalized to zero mean and unit variance. In training, random patches of size $96 \times 96 \times 96$ were used, and the same size was applied with a stride of $16\times16\times4$ in inference. \textbf{For BraTS2019 dataset}, we used 250 images for training, 60 images for testing and 25 images for validation as reported in \cite{CPCL}. All images were resampled to $1 mm^{3}$ resolution and normalized. Images were randomly cropped into patches of size $96 \times 96 \times 96$ in training, and the same size with a stride of $64 \times 64 \times 64$ in inference.

2) Evaluation Metrics: We evaluated our method on four metrics, including Dice Similarity Coefficient (Dice [\%]), Jaccard Index (Jaccard [\%]), 95\% Hausdorff Distance (95HD [voxel]), and Average Surface Distance (ASD [voxel]). (1) Overall performance: Dice and Jaccard quantify the overall segmentation performance by measuring volumetric overlap between predictions and ground truth, where higher values indicate better performance. While both assess region-level agreement, Dice is more sensitive to larger regions, whereas Jaccard provides a stricter penalty for misclassifications. (2) Boundary performance: 95HD and ASD evaluate boundary localization accuracy through surface distance measurements, where lower values indicate better boundary precision. 95HD captures the maximum surface error at the 95th percentile, making it robust to outliers, while ASD reflects the average precision of boundary delineation across the entire contour.

\subsection{Comparison with other SOTA methods}
To demonstrate the effectiveness of our proposed method, we compared it with the following methods:

(1) Supervised-only Segmentation methods(SS): V-Net \cite{vnetdiceloss} trained with 100\% labeled data establishes the upper bound of performance, while V-Net trained with only 5\% and 10\% labeled data sets the lower bound, illustrating the performance range under varying supervision levels.

(2) Consistency Regularization-based (CR) methods: We compared against uncertainty-aware mean teacher (UA-MT) \cite{UAMT}, dual-task consistency (DTC) \cite{DTC}, and triple-task mutual consistency (TTMC) \cite{TTMC} to evaluate the effectiveness of different consistency constraint strategies.

(3) Single Prototype Learning-based (SPL) methods: Cyclic Prototype Consistency Learning (CPCL) \cite{CPCL} and Uncertainty-informed Prototype COnsistency Learning (UPCoL) \cite{upcol} are included to demonstrate the representational limitations of single prototype methods.

(4) Multiple Prototype Learning-based (MPL) methods: Boundary-aware prototype consistency (BaPC) \cite{bapc} and multi-prototype embedding refinement (MPER) \cite{mper} are compared to validate the superior performance of our method in capturing more diverse representations and more precise segmentation results. 

To ensure fair comparison, all methods utilized identical VNet \cite{vnetdiceloss} backbone architecture, data preprocessing and data partition. We employed official codes and reported the corresponding results for each method. Since BaPC does not provide open-source code, we reproduce it based on the original paper and report the results. All results are reported as mean$\pm$standard deviation. Statistical significance is assessed using a two-sided paired t-test comparing MPCL against each  method, with p$\leq$0.05 indicated by * in the tables.

\subsubsection{Quantitative Evaluation for Metric Superiority}
\begin{figure}
    \centering
    \includegraphics[width=1\linewidth]{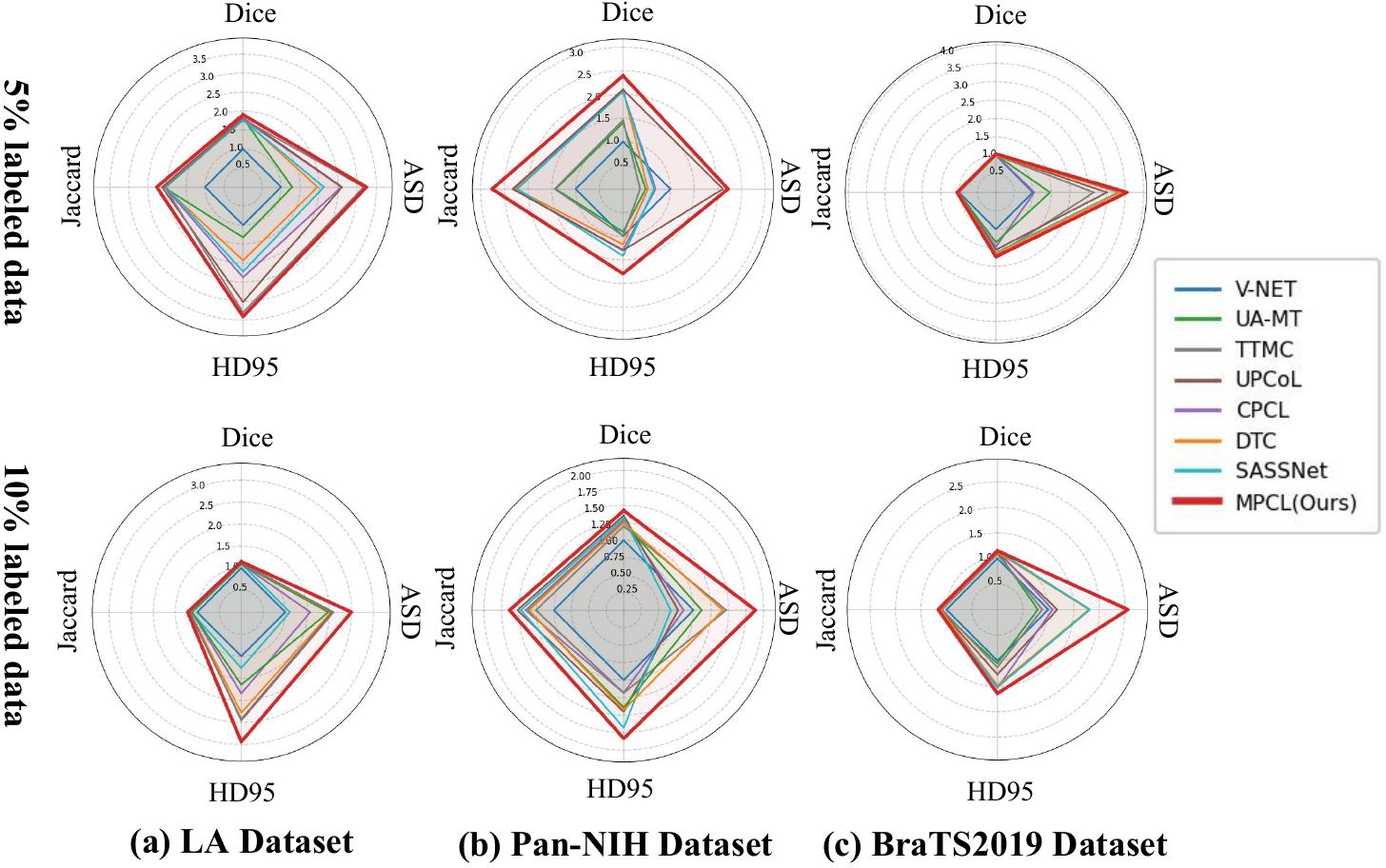}
    \caption{Radar maps comparison shows the superiority of our MPCL in all metrics, especially in boundary precision, on LA, Pan-NIH and BraTS2019 dataset under 5\% and 10\% labeled data setting. All metrics are normalized with respect to the baseline V-NET.}
    \label{radarfigure}
\end{figure}

Tables \ref{ladataset}, \ref{pandataset}, and \ref{bratsdataset} present a quantitative comparison between our method (MPCL) and other methods on the LA, Pan-NIH, and BraTS2019 datasets under 5\% and 10\% labeled data settings. Our method attains the best performance across all metrics under both settings, achieving precise segmentation results and demonstrating substantial improvements over other state-of-the-art methods. The radar maps in Fig. \ref{radarfigure} further visualize our method's significant improvements across all four evaluation metrics on each dataset especially in boundary performance. It can be observed that: \textbf{(1)} Compared with SS baselines, all semi-supervised methods demonstrate significant improvements, confirming the benefit of incorporating unlabeled data. Notably, as the amount of labeled data decreases, SS methods degrade rapidly because they lack additional information beyond the limited annotations. In contrast, semi-supervised methods leverage unlabeled data to provide more training samples, and different consistency regularization strategies further enhance the reliable utilization of unlabeled data. Our MPCL achieves remarkable improvements of nearly 52\% on LA dataset and 34\% on Pan-NIH dataset under 5\% labeled data setting compared to the SS baseline. \textbf{(2)} Compared with CR methods SPL methods achieve competitive and even superior results under 10\% labeled data setting, benefiting from efficient prototype-based representations that effectively capture class-level characteristics. However, when labeled data decreases to 5\%, SPL methods exhibit equal or degraded performance. This is attributed to the lack of diversity in representations. Relying on a single prototype per class yields uniform representations that fail to capture diverse structural heterogeneity, particularly limiting effectiveness under extremely sparse supervision. \textbf{(3)} MPL methods consistently outperform both SPL and CR methods across most settings. This superiority stems from the fact that multiple prototypes provide richer and more diverse representations, which becomes especially crucial when labeled data is limited. With multiple prototypes, the model can better capture intra-class variations and structural complexity, leading to more robust feature learning and improved segmentation precision.

Compared with other MPL methods, our method achieves the best performance and demonstrates superior precision when labeled data becomes scarce. This better precision is particularly evident on the challenging Pan-NIH dataset, as also clearly visualized in the radar comparison (Fig. \ref{radarfigure}): when labeled data decreases from 10\% to 5\%, BaPC experiences a 33.51\% Dice drop with 39.04 voxels increase in 95HD, and MPER suffers a 26.64\% Dice drop with 30.87 voxels increase in 95HD. In contrast, our MPCL maintains relative stability with only 17.45\% Dice decrease and 13.92 voxels increase in 95HD. This superior performance can be attributed to several key factors. 
First, BaPC and MPER lack explicit guidance to capture intensity-based heterogeneity. Without such guidance, the network gravitates toward dominant features, causing multiple prototypes to concentrate on similar patterns rather than diverse structural variations, essentially suffering from uniform representation as SPL methods despite using multiple prototypes.
Second, the prototypes in these methods passively rely on the segmentation network's learned representations without being independently refined at the representation level, which limits both the flexibility and effectiveness of prototype optimization.
% Second, these methods perform pixel-level alignment between prototypes and image features to optimize the prototypical space, which introduces information redundancy and makes optimization less effective at the representation level.
In contrast, our IHPG explicitly models intra-class heterogeneity by leveraging intensity characteristics, capturing not only the heterogeneity within the target organ but also the heterogeneity in background regions. This intensity-based intra-class heterogeneity-aware approach provides more diverse and clinically meaningful representations of image features. 
Furthermore, our PSO operates at the prototype level, significantly enhancing inter-class separability and creating a more discriminative prototypical space. With the application of DKA, the segmentation network not only learns intensity-based intra-class heterogeneity from the prototype learning branch, but also provides more stable feature representations back to the prototype learning branch, forming a bidirectional information flow that mutually reinforces both branches and ultimately yields more precise segmentation results.

\begin{table}[]
\caption{Comparison on LA Dataset under 5\% and 10\% labeled data setting demonstrates the superiority of our MPCL. The best results are highlighted in bold. 'Lb' and 'ULb' denote labeled and unlabeled number.}
\label{ladataset}
\centering
\renewcommand{\arraystretch}{1.2}
\resizebox{\linewidth}{!}{
\begin{tabular}{c|c|c|cccc}
\bottomrule[1pt]
\textbf{Method}           & Type & \textbf{Lb/ULb}             & \textbf{Dice↑} & \textbf{Jaccard↑} & \textbf{HD95↓} & \textbf{ASD↓} \\ \hline
V-NET \cite{vnetdiceloss} & SS   & 80 (100\%)                & 91.33$_{\pm 1.95}$ & 84.10$_{\pm 3.30}$ & 5.60$_{\pm 2.26}$ & 1.48$_{\pm 0.44}$ \\
V-NET \cite{vnetdiceloss} & SS   & 4/0 (5\%)                   & 44.98$_{\pm 27.65}$* & 33.27$_{\pm 24.13}$* & 36.59$_{\pm 24.87}$* & 10.12$_{\pm 8.45}$* \\
V-NET \cite{vnetdiceloss} & SS   & 8/0 (10\%)                  & 78.88$_{\pm 19.27}$* & 68.20$_{\pm 20.53}$* & 14.92$_{\pm 12.20}$* & 2.60$_{\pm 1.34}$* \\ \hline
UA-MT \cite{UAMT}         & CR   & \multirow{8}{*}{4/76 (5\%)} & 81.55$_{\pm 7.34}$* & 69.44$_{\pm 9.76}$* & 27.57$_{\pm 16.57}$* & 7.83$_{\pm 4.81}$* \\
DTC \cite{DTC}            & CR   &                             & 80.56$_{\pm 10.70}$* & 68.63$_{\pm 13.27}$* & 18.94$_{\pm 15.58}$* & 5.17$_{\pm 4.98}$* \\
TTMC \cite{TTMC}          & CR   &                             & 80.33$_{\pm 9.69}$* & 68.18$_{\pm 12.99}$* & 11.61$_{\pm 5.54}$* & 3.14$_{\pm 1.68}$* \\
CPCL \cite{CPCL}          & SPL  &                             & 80.65$_{\pm 9.09}$* & 68.46$_{\pm 11.73}$* & 15.38$_{\pm 9.24}$* & 3.92$_{\pm 2.70}$* \\
UPCoL \cite{upcol}        & SPL  &                             & 82.02$_{\pm 11.22}$* & 70.88$_{\pm 14.36}$* & 12.08$_{\pm 10.44}$* & 3.89$_{\pm 4.37}$ \\
BaPC \cite{bapc}          & MPL  &                             & 84.53$_{\pm 6.10}$* & 73.68$_{\pm 8.53}$* & 12.07$_{\pm 7.33}$* & 3.07$_{\pm 1.52}$* \\
MPER \cite{mper}          & MPL  &                             & 86.07$_{\pm 3.28}$ & 75.69$_{\pm 5.00}$ & 11.19$_{\pm 6.05}$ & 2.96$_{\pm 2.05}$ \\
\textbf{MPCL(Ours)}       & MPL  &                             & \textbf{86.91}$_{\pm 4.74}$ & \textbf{77.15}$_{\pm 7.22}$ & \textbf{7.97}$_{\pm 6.54}$ & \textbf{2.48}$_{\pm 1.15}$ \\ \hline
UA-MT \cite{UAMT}         & CR   & \multirow{8}{*}{8/72 (10\%)} & 86.08$_{\pm 4.85}$* & 75.85$_{\pm 7.22}$* & 11.21$_{\pm 5.44}$* & 2.29$_{\pm 1.02}$* \\
DTC \cite{DTC}            & CR   &                             & 88.13$_{\pm 3.60}$* & 78.95$_{\pm 5.71}$* & 8.09$_{\pm 4.24}$* & 2.22$_{\pm 1.04}$* \\
TTMC \cite{TTMC}          & CR   &                             & 87.51$_{\pm 7.14}$* & 78.40$_{\pm 10.02}$* & 7.50$_{\pm 5.45}$* & 2.16$_{\pm 1.23}$* \\
CPCL \cite{CPCL}          & SPL  &                             & 86.06$_{\pm 5.35}$* & 75.88$_{\pm 7.99}$* & 10.00$_{\pm 6.15}$* & 2.89$_{\pm 1.61}$* \\
UPCoL \cite{upcol}        & SPL  &                             & 88.60$_{\pm 4.98}$* & 79.85$_{\pm 7.55}$* & 7.59$_{\pm 5.69}$ & 2.19$_{\pm 1.47}$ \\
BaPC \cite{bapc}          & MPL  &                             & 88.66$_{\pm 4.31}$* & 79.89$_{\pm 6.64}$* & 7.07$_{\pm 4.38}$ & 1.97$_{\pm 1.07}$ \\
MPER \cite{mper}          & MPL  &                             & 89.15$_{\pm 2.99}$ & 80.55$_{\pm 4.81}$ & 7.70$_{\pm 2.98}$* & 1.90$_{\pm 0.71}$ \\
\textbf{MPCL(Ours)}       & MPL  &                             & \textbf{90.35}$_{\pm 2.70}$ & \textbf{82.50}$_{\pm 4.42}$ & \textbf{6.28}$_{\pm 3.85}$ & \textbf{1.81}$_{\pm 0.85}$ \\  \toprule[1pt]
\end{tabular}}
\end{table}

\begin{table}[]
\caption{Comparison on Pan-NIH Dataset under 5\% and 10\% labeled data setting demonstrates the superiority of our MPCL. The best results are highlighted in bold. 'Lb' and 'ULb' denote labeled and unlabeled number.}
\label{pandataset}
\centering
\renewcommand{\arraystretch}{1.2}
\resizebox{\linewidth}{!}{
\begin{tabular}{c|c|c|cccc}
\bottomrule[1pt]
\textbf{Method}           & Type & \textbf{Lb/ULb}             & \textbf{Dice↑} & \textbf{Jaccard↑} & \textbf{HD95↓} & \textbf{ASD↓} \\ \hline
V-NET \cite{vnetdiceloss} & SS   & 60/0 (100\%)                & 83.11$_{\pm 5.57}$ & 71.46$_{\pm 7.85}$ & 5.16$_{\pm 3.76}$ & 1.05$_{\pm 0.29}$ \\
V-NET \cite{vnetdiceloss} & SS   & 3/0 (5\%)                   & 24.64$_{\pm 21.18}$* & 15.84$_{\pm 14.91}$* & 42.74$_{\pm 42.43}$ & 7.00$_{\pm 6.22}$* \\
V-NET \cite{vnetdiceloss} & SS   & 6/0 (10\%)                  & 53.76$_{\pm 17.25}$* & 38.58$_{\pm 16.25}$* & 20.35$_{\pm 15.05}$* & 3.17$_{\pm 5.35}$ \\ \hline
UA-MT \cite{UAMT}         & CR   & \multirow{8}{*}{3/57 (5\%)}  & 35.71$_{\pm 12.21}$* & 22.41$_{\pm 9.08}$* & 57.97$_{\pm 13.81}$* & 25.12$_{\pm 7.22}$* \\
DTC \cite{DTC}            & CR   &                             & 51.90$_{\pm 11.91}$ & 35.95$_{\pm 11.32}$ & 36.42$_{\pm 14.64}$* & 13.35$_{\pm 5.66}$* \\
TTMC \cite{TTMC}          & CR   &                             & 35.90$_{\pm 14.42}$* & 22.83$_{\pm 10.90}$* & 43.32$_{\pm 12.83}$* & 19.34$_{\pm 8.01}$* \\
CPCL \cite{CPCL}          & SPL  &                             & 50.88$_{\pm 21.96}$ & 36.72$_{\pm 17.84}$ & 33.16$_{\pm 23.76}$ & 10.19$_{\pm 14.51}$ \\
UPCoL \cite{upcol}        & SPL  &                             & 51.77$_{\pm 18.84}$* & 36.86$_{\pm 15.41}$* & 32.90$_{\pm 24.42}$ & 3.32$_{\pm 3.03}$ \\
BaPC \cite{bapc}          & MPL  &                             & 41.59$_{\pm 18.19}$* & 27.88$_{\pm 14.30}$* & 51.30$_{\pm 30.85}$* & 3.86$_{\pm 5.58}$* \\
MPER \cite{mper}          & MPL  &                             & 49.23$_{\pm 23.31}$ & 35.53$_{\pm 18.80}$ & 35.47$_{\pm 23.49}$ & 6.61$_{\pm 11.08}$ \\
\textbf{MPCL(Ours)}       & MPL  &                             & \textbf{59.01}$_{\pm 17.77}$ & \textbf{43.72}$_{\pm 14.95}$ & \textbf{23.85}$_{\pm 17.55}$ & \textbf{3.16}$_{\pm 3.84}$ \\ \hline
UA-MT \cite{UAMT}         & CR   & \multirow{8}{*}{6/54 (10\%)} & 67.66$_{\pm 15.16}$* & 52.78$_{\pm 15.35}$* & 14.64$_{\pm 12.25}$ & 2.84$_{\pm 2.03}$* \\
DTC \cite{DTC}            & CR   &                             & 67.32$_{\pm 14.93}$* & 52.34$_{\pm 15.14}$* & 14.27$_{\pm 14.22}$ & 2.23$_{\pm 1.66}$ \\
TTMC \cite{TTMC}          & CR   &                             & 64.56$_{\pm 10.73}$* & 48.52$_{\pm 11.39}$* & 18.99$_{\pm 16.59}$* & 5.83$_{\pm 4.92}$* \\
CPCL \cite{CPCL}          & SPL  &                             & 69.09$_{\pm 13.89}$* & 54.23$_{\pm 14.60}$* & 17.19$_{\pm 15.45}$* & 3.69$_{\pm 3.76}$* \\
UPCoL \cite{upcol}        & SPL  &                             & 72.78$_{\pm 13.80}$* & 58.72$_{\pm 14.86}$* & 14.00$_{\pm 15.97}$ & 4.07$_{\pm 3.98}$* \\
BaPC \cite{bapc}          & MPL  &                             & 75.10$_{\pm 11.18}$* & 61.24$_{\pm 12.50}$ & 12.26$_{\pm 15.82}$* & 3.26$_{\pm 4.49}$ \\
MPER \cite{mper}          & MPL  &                             & 73.56$_{\pm 9.29}$ & 58.99$_{\pm 11.14}$* & 10.13$_{\pm 7.19}$ & 2.30$_{\pm 2.30}$* \\
\textbf{MPCL(Ours)}       & MPL  &                             & \textbf{76.46}$_{\pm 9.99}$ & \textbf{62.81}$_{\pm 11.48}$ & \textbf{9.93}$_{\pm 14.33}$ & \textbf{1.80}$_{\pm 0.85}$  \\ \toprule[1pt]
\end{tabular}}
\end{table}

\begin{table}[]
\caption{Comparison on BraTS2019 Dataset under two labeled data setting  demonstrates the superiority of our MPCL. The best results are highlighted in bold. 'Lb' and 'ULb' denote labeled and unlabeled number.}
\label{bratsdataset}
\centering
\renewcommand{\arraystretch}{1.2}
\resizebox{\linewidth}{!}{
\begin{tabular}{c|c|c|cccc}
\bottomrule[1pt]
\textbf{Method}           & Type & \textbf{Lb/ULb}                & \textbf{Dice↑} & \textbf{Jaccard↑} & \textbf{HD95↓} & \textbf{ASD↓} \\ \hline
V-NET \cite{vnetdiceloss} & SS   & 250/0 (100\%)                  & 86.08$_{\pm 9.82}$ & 76.70$_{\pm 13.53}$ & 9.27$_{\pm 14.64}$ & 2.53$_{\pm 3.93}$ \\
V-NET \cite{vnetdiceloss} & SS   & 13/0 (5\%)                     & 71.40$_{\pm 29.12}$* & 62.03$_{\pm 29.09}$* & 16.62$_{\pm 18.57}$ & 3.69$_{\pm 5.65}$* \\
V-NET \cite{vnetdiceloss} & SS   & 25/0 (10\%)                    & 76.94$_{\pm 20.44}$* & 66.18$_{\pm 22.71}$* & 13.41$_{\pm 15.45}$* & 2.06$_{\pm 2.54}$* \\ \hline
UA-MT \cite{UAMT}         & CR   & \multirow{8}{*}{13/237 (5\%)}  & 77.73$_{\pm 20.92}$ & 67.41$_{\pm 22.82}$ & 15.78$_{\pm 18.89}$ & 4.45$_{\pm 7.04}$* \\
DTC \cite{DTC}            & CR   &                                & 76.88$_{\pm 22.69}$ & 66.87$_{\pm 24.43}$ & 12.83$_{\pm 14.47}$ & 3.22$_{\pm 3.46}$ \\
TTMC \cite{TTMC}          & CR   &                                & 76.89$_{\pm 23.51}$ & 67.09$_{\pm 24.73}$ & 12.50$_{\pm 15.51}$ & 2.46$_{\pm 5.33}$ \\
CPCL \cite{CPCL}          & SPL  &                                & 77.62$_{\pm 21.96}$ & 67.60$_{\pm 23.66}$ & 14.52$_{\pm 21.09}$ & 6.26$_{\pm 14.61}$* \\
UPCoL \cite{upcol}        & SPL  &                                & 77.69$_{\pm 18.94}$* & 66.89$_{\pm 21.98}$* & 13.73$_{\pm 14.94}$* & 2.17$_{\pm 2.50}$* \\
BaPC \cite{bapc}          & MPL  &                                & 78.17$_{\pm 22.63}$ & 68.59$_{\pm 24.18}$ & 13.32$_{\pm 15.90}$ & 1.99$_{\pm 2.75}$ \\
MPER \cite{mper}          & MPL  &                                & 75.65$_{\pm 23.26}$* & 65.48$_{\pm 25.21}$* & 14.88$_{\pm 16.70}$* & 3.65$_{\pm 4.26}$* \\
\textbf{MPCL(Ours)}       & MPL  &                                & \textbf{80.10}$_{\pm 15.87}$ & \textbf{69.35}$_{\pm 19.33}$ & \textbf{11.86}$_{\pm 13.63}$ & \textbf{1.97}$_{\pm 2.33}$ \\ \hline
UA-MT \cite{UAMT}         & CR   & \multirow{8}{*}{25/225 (10\%)} & 80.84$_{\pm 16.40}$ & 70.44$_{\pm 19.39}$ & 15.73$_{\pm 20.38}$* & 4.65$_{\pm 6.13}$* \\
DTC \cite{DTC}            & CR   &                                & 79.99$_{\pm 16.12}$* & 69.26$_{\pm 19.90}$* & 11.19$_{\pm 13.27}$ & 2.05$_{\pm 2.72}$* \\
TTMC \cite{TTMC}          & CR   &                                & 77.61$_{\pm 21.90}$* & 67.55$_{\pm 23.84}$* & 14.70$_{\pm 20.61}$* & 4.19$_{\pm 12.95}$ \\
CPCL \cite{CPCL}          & SPL  &                                & 82.60$_{\pm 14.65}$ & 72.61$_{\pm 18.53}$ & 10.82$_{\pm 13.94}$ & 3.41$_{\pm 5.05}$* \\
UPCoL \cite{upcol}        & SPL  &                                & 80.21$_{\pm 18.37}$* & 70.08$_{\pm 21.07}$* & 13.07$_{\pm 18.50}$ & 3.14$_{\pm 11.30}$ \\
BaPC \cite{bapc}          & MPL  &                                & 80.22$_{\pm 17.26}$ & 69.91$_{\pm 20.65}$* & 10.76$_{\pm 12.63}$ & 2.29$_{\pm 2.99}$* \\
MPER \cite{mper}          & MPL  &                                & 81.53$_{\pm 14.55}$ & 71.11$_{\pm 18.74}$ & 10.29$_{\pm 12.39}$ & 2.53$_{\pm 2.62}$* \\
\textbf{MPCL(Ours)}       & MPL  &                                & \textbf{83.40}$_{\pm 13.23}$ & \textbf{73.45}$_{\pm 17.12}$ & \textbf{9.83}$_{\pm 13.17}$ & \textbf{1.70}$_{\pm 2.25}$ \\ \toprule[1pt]
\end{tabular}}
\end{table}

\subsubsection{Qualitative Evaluation for Visual Superiority}
\begin{figure*}[t]
    \centering
    \includegraphics[width=1\textwidth]{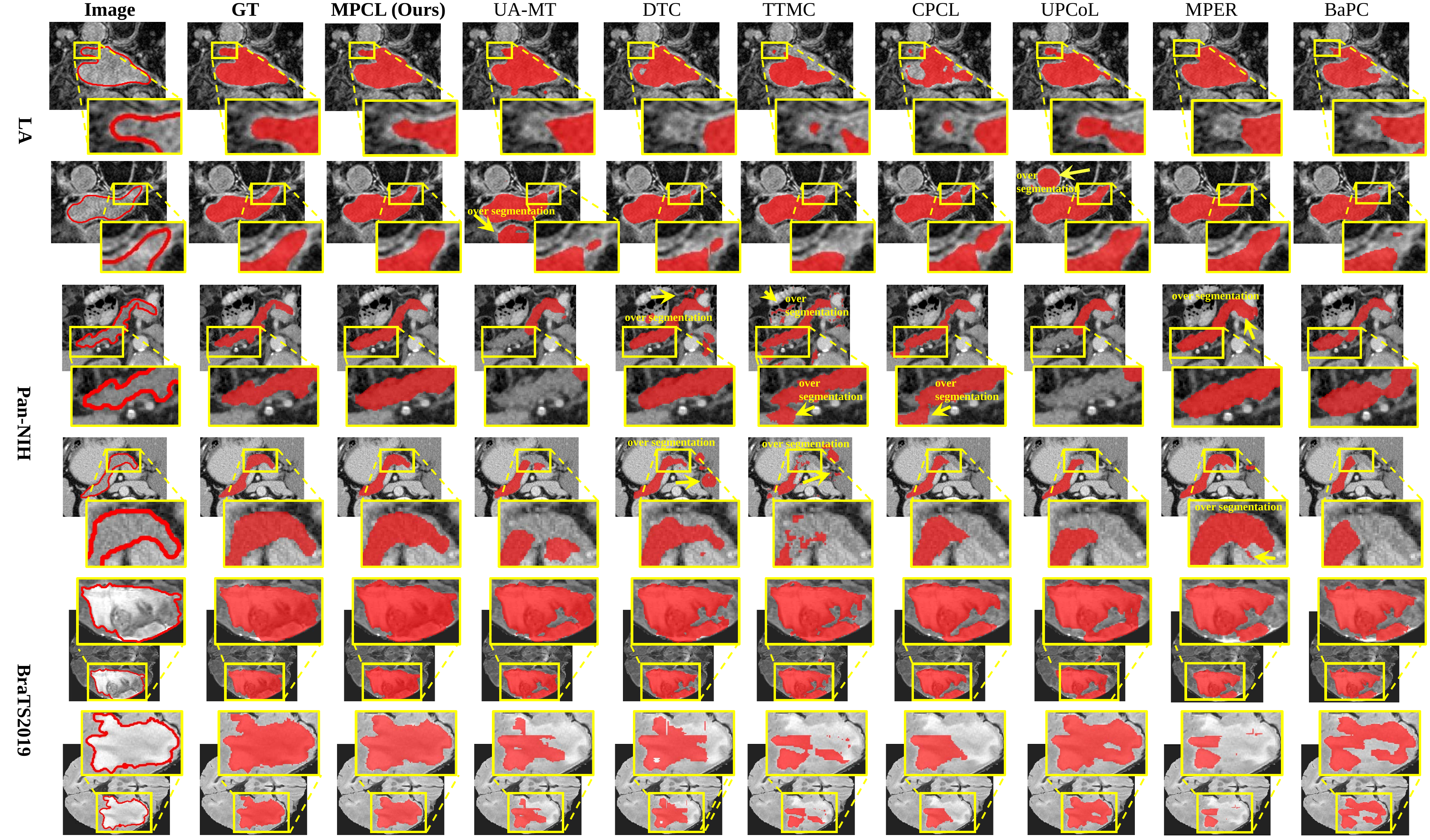}
    \caption{Visualization comparison indicates more precise segmentation of our MPCL on examples of 2D slices on LA, Pan-NIH and BraTS2019 dataset under 5\% labeled data setting. Most other methods struggle to process heterogeneous foreground and to distinguish similar foreground and background, thus causing under-segmentation in the foreground and over-segmentation (yellow arrow) in the background.}
    \label{expcomp}
    % \vspace{-15pt}
\end{figure*}

Fig. \ref{expcomp} provides 2D slice visual comparisons under 5\% labeled data setting across all three datasets, demonstrating that our method achieves segmentation results with superior precision.
Two key observations can be made: (1) Most competing methods struggle to segment heterogeneous foreground regions, resulting in under-segmentation. For instance, in the LA dataset, sharp apical structures exhibit distinct intensity characteristics—either dark (LA example in the first row) or light (LA example in the second row)—while the main body of the left atrium is mainly gray. Consequently, most methods fail to adequately segment these apical regions due to their intensity deviation from the dominant pattern. Similarly, in the BraTS2019 dataset, the tumor regions clearly exhibit distinct intensity distributions (light, gray, and dark). However, as shown in the first and second rows of BraTS2019, all methods only capture the light and gray parts while missing the dark parts. This is because light and gray parts represent dominant tumor features, while dark parts are also prevalent in the background, leading methods to misclassify dark tumor regions as background. In contrast, our method explicitly follows intensity characteristics to achieve more precise foreground segmentation, capturing all intensity-based parts within the target structure.
(2) Some methods struggle to distinguish similar foreground and background regions, causing over-segmentation in the background (indicated by yellow arrows). In both LA and Pan-NIH datasets, the segmentation targets share strong intensity similarities with surrounding tissues, making discrimination challenging. This is particularly evident in the first row of Pan-NIH examples, where the background tissue near the pancreatic tail exhibits intensity characteristics highly similar to the target organ. Methods such as TTMC and CPCL consequently over-segment these background regions. In contrast, our method enhances inter-class discriminability in the prototypical space, creating greater separation between similar foreground and background regions, thereby achieving more precise segmentation.
For MPL methods, both MPER and BaPC lack the guidance of intensity-aligned prototype generation. Without explicit intensity-based guidance, these methods exhibit under-segmentation in anatomical structures with substantial intra-class heterogeneity and struggle to distinguish foreground regions that share similar intensities with the background.

\subsection{Ablation studies}
\subsubsection{Effectiveness of Key Designs}
To validate the effectiveness of each key design in our proposed MPCL, we conducted a comprehensive ablation study on the LA dataset under 10\% labeled data setting, as shown in Table \ref{ablation}.
We establish our baseline using VNet, which is supervised only on labeled data. Subsequently, we incorporate the Mean-Teacher (MT) architecture to leverage unlabeled data through segmentation prediction consistency, which substantially improves performance over the vanilla VNet. The Average over Multiple Classifiers (AMC) strategy further enhances results, and is adopted in all subsequent experiments. 
We further introduce the prototype learning branch to construct the SPL baseline, where a single prototype per class is generated through Masked Average Pooling (MAP) and transferred to the segmentation network via DKA, yielding additional performance gains and achieving 88.60\% Dice (+1.24\%).
This validates the effectiveness of SPL baseline as a foundation for capturing more precise segmentation results. 
To evaluate the benefit of capturing intra-class heterogeneity in IHPG, we compare the performance of SPL(only DKA) and MPL(IHPG+DKA) configurations. These multiple prototypes are generated aligned with intensity characteristics in anatomical structures, providing more diverse representations. The integration of IHPG improves performance to 89.42\% Dice, with notable improvements in boundary metrics. This demonstrates that capturing intra-class heterogeneity through multiple prototypes leads to more diverse structural representations. 
The design of PSO enhances the discriminative power of prototypes, creating a more robust and generalizable prototypical space. The advantages brought by PSO in both SPL and MPL are noteworthy. Notably, the configuration of PSO in SPL baseline setting (PSO+DKA) with 89.63\% Dice performs better than the MPL baseline configuration alone (IHPG+DKA) with 89.42\% Dice.  This demonstrates that PSO builds more discriminative representations in the prototypical space and is capable of better distinguishing differences between classes.
The integration of all three designs (IHPG+PSO+DKA) achieves the best performance across all metrics, representing substantial improvements of +2.99\% Dice over the MT+AMC baseline and +1.75\% Dice over the simple SPL configuration. This further demonstrates the effectiveness of our proposed method in providing more diverse representations and more precise segmentation results in SSMIS.

\begin{table}[h]
% \color{blue}
\caption{Ablation Study for Key Designs demonstrates the effectiveness of our three designs in MPCL on LA Dataset under 10\% labeled data setting. The best results are highlighted in bold.}
\label{ablation}
\centering
\renewcommand{\arraystretch}{1.2}
\resizebox{\linewidth}{!}{
\begin{tabular}{l|ccc|cccc}
\toprule[1pt]
\multirow{2}{*}{\textbf{Baseline}} & \multicolumn{3}{c|}{\textbf{Components (Ours)}} & \multirow{2}{*}{\textbf{Dice$\uparrow$}} & \multirow{2}{*}{\textbf{Jaccard$\uparrow$}} & \multirow{2}{*}{\textbf{95HD$\downarrow$}} & \multirow{2}{*}{\textbf{ASD$\downarrow$}} \\
\cmidrule(lr){2-4}
& \textbf{IHPG} & \textbf{PSO} & \textbf{DKA} & & & & \\
\midrule
VNet  & & & & 78.77 & 67.79 & 18.47 & 4.53 \\
\midrule
MT & & & & 85.73 & 75.47 & 11.57 & 2.85 \\
+AMC & & & & 87.36 & 77.79 & 8.68 & 2.14 \\
\midrule
\multirow{2}{*}{SPL} & & & $\checkmark$ & 88.60 & 79.85 & 7.59 & 2.19 \\
 & & $\checkmark$ & $\checkmark$ & 89.63 & 81.37 & 7.06 & 1.84 \\
\midrule
\multirow{2}{*}{MPL} & $\checkmark$ & & $\checkmark$ & 89.42 & 81.15 & 7.20 & 1.86 \\
 & $\checkmark$ & $\checkmark$ & $\checkmark$ & \textbf{90.35} & \textbf{82.50} & \textbf{6.28} & \textbf{1.81} \\
\bottomrule[1pt]
\end{tabular}}
\end{table}

\begin{table}[]
\caption{Ablation Study for Impact of Prototype Number on three datasets under 5\% labeled data setting. The best results are highlighted in bold.}
\label{ablationnumber}
\centering
\renewcommand{\arraystretch}{1.2}
\resizebox{0.8\linewidth}{!}{
\begin{tabular}{c|c|cccc}
\bottomrule[1pt]
\textbf{Dataset}         & \textbf{K} & \textbf{Dice↑} & \textbf{Jaccard↑} & \textbf{95HD↓} & \textbf{ASD↓} \\ \hline
\multirow{5}{*}{LA}      & 1          & 83.34          & 72.34             & 12.71          & 3.62          \\
                         & 2          & 84.83          & 74.38             & 11.49          & 3.17          \\
                         & 3          & \textbf{86.91} & \textbf{77.15}    & \textbf{7.97} & \textbf{2.48} \\
                         & 4          &85.02  &74.66   & 11.67 & 3.27\\
                         & 5          &85.31  & 75.01  &11.90  &3.28 \\\hline
\multirow{5}{*}{Pan-NIH} & 1          & 51.42          & 37.89             & 26.91          & 5.34          \\
                         & 2          & 57.60          & 41.85             & 30.48          & 3.40          \\
                         & 3          & \textbf{59.01} & \textbf{43.72}    & \textbf{23.85} & 3.16 \\
                         & 4          &58.46  &43.10   &24.17  &\textbf{2.83}  \\
                         & 5          &58.00  &42.64   &28.11  &3.68 \\
\hline
\multirow{5}{*}{BraTS2019} & 1          & 77.87          & 67.12             & 12.73          & 2.01          \\
                         & 2          & 79.62          & 68.89             & \textbf{11.79}          & \textbf{1.72}          \\
                         & 3          & \textbf{80.10} & \textbf{69.35}    & 11.86 & 1.97 \\
                         & 4          &79.59  &68.82   &13.01  &2.02  \\
                         & 5          &79.92  &69.09   &11.91  &2.10 \\
                         \toprule[1pt]
\end{tabular}}
\end{table}

\subsubsection{Analysis of Prototype Number}
As defined in Sec. \ref{sec:ihpg}, $K$ defines the number of prototypes. We investigated the impact of prototype number $K$  varying from 1 to 5 under a more challenging 5\% labeled data setting on LA, Pan-NIH and BraTS2019 datasets, as shown in Table \ref{ablationnumber}.
It can be observed that the performance consistently improves as the prototype number increases from $K=1$ to $K=3$ across all three datasets. On the LA dataset, Dice improves by +3.57\% and 95HD reduces by 1.14 voxels. The improvement is particularly pronounced on the challenging Pan-NIH dataset, achieving a substantial +7.59\% Dice improvement. For BraTS2019, which exhibits substantial intra-class heterogeneity in tumor regions, increasing from $K=1$ to $K=3$ prototypes improves Dice by +2.23\%. However, further increasing the prototype number from $K=3$ to $K=5$ yields diminishing returns, with performance plateauing or even slightly declining on some datasets, while computational costs grow correspondingly. Notably, the optimal selection of 3 prototypes aligns with our intrinsic intensity characteristics that anatomical structures can be naturally partitioned into three distinct intensity-based subregions (light, gray, and dark). This validates that IHPG effectively captures the intrinsic intra-class heterogeneity through intensity characteristics and provides more diverse structural representations. Therefore, considering the balance between performance, computational efficiency, and the alignment with intensity-based intra-class heterogeneity, we select $K=3$ for our final experiments.

\begin{figure}
    \centering
    \includegraphics[width=1\linewidth]{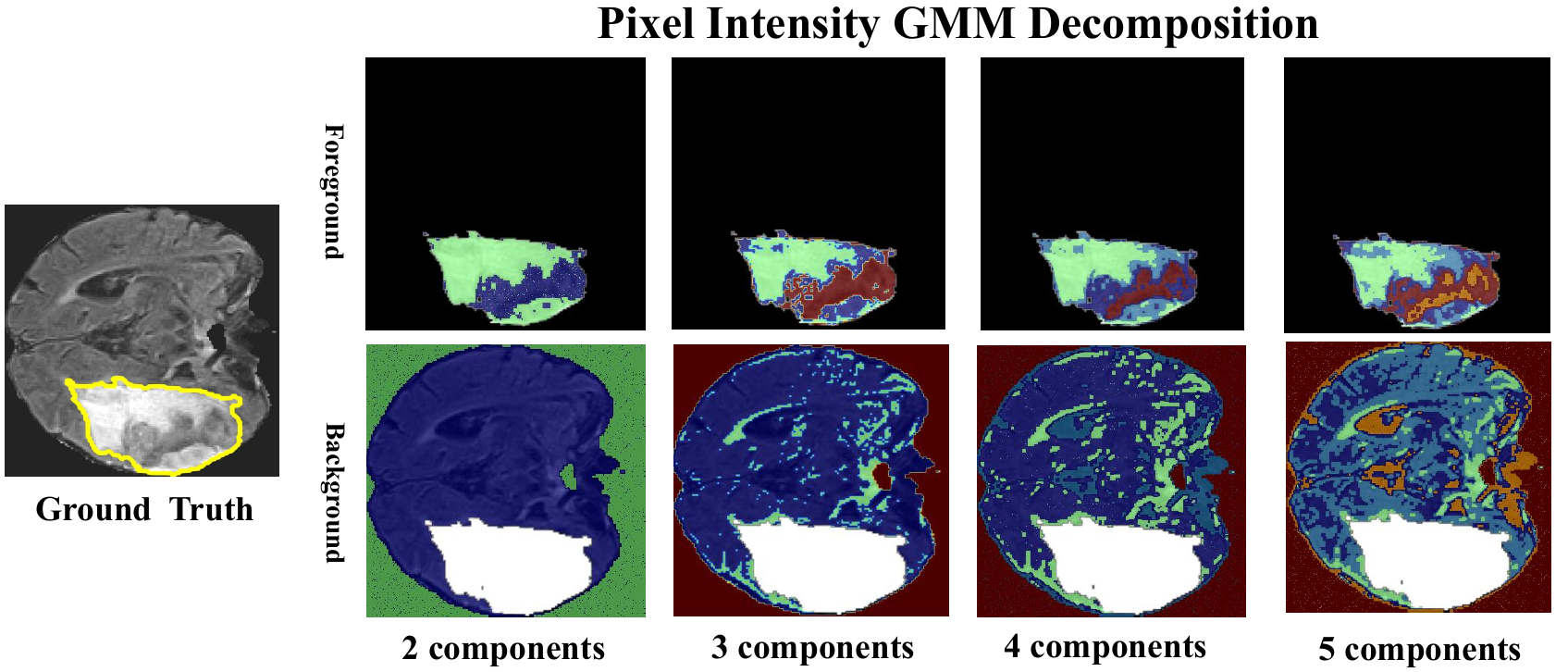}
    \caption{GMM decomposition of the intensity distribution under different component number settings reveals that increasing the component number beyond 3 leads to redundant subdivision within existing intensity subregions, where individual components lose their representativeness of distinct intensity-based patterns.}
    \label{GMMnumber}
\end{figure}

\begin{figure}
    \centering
    \includegraphics[width=0.9\linewidth]{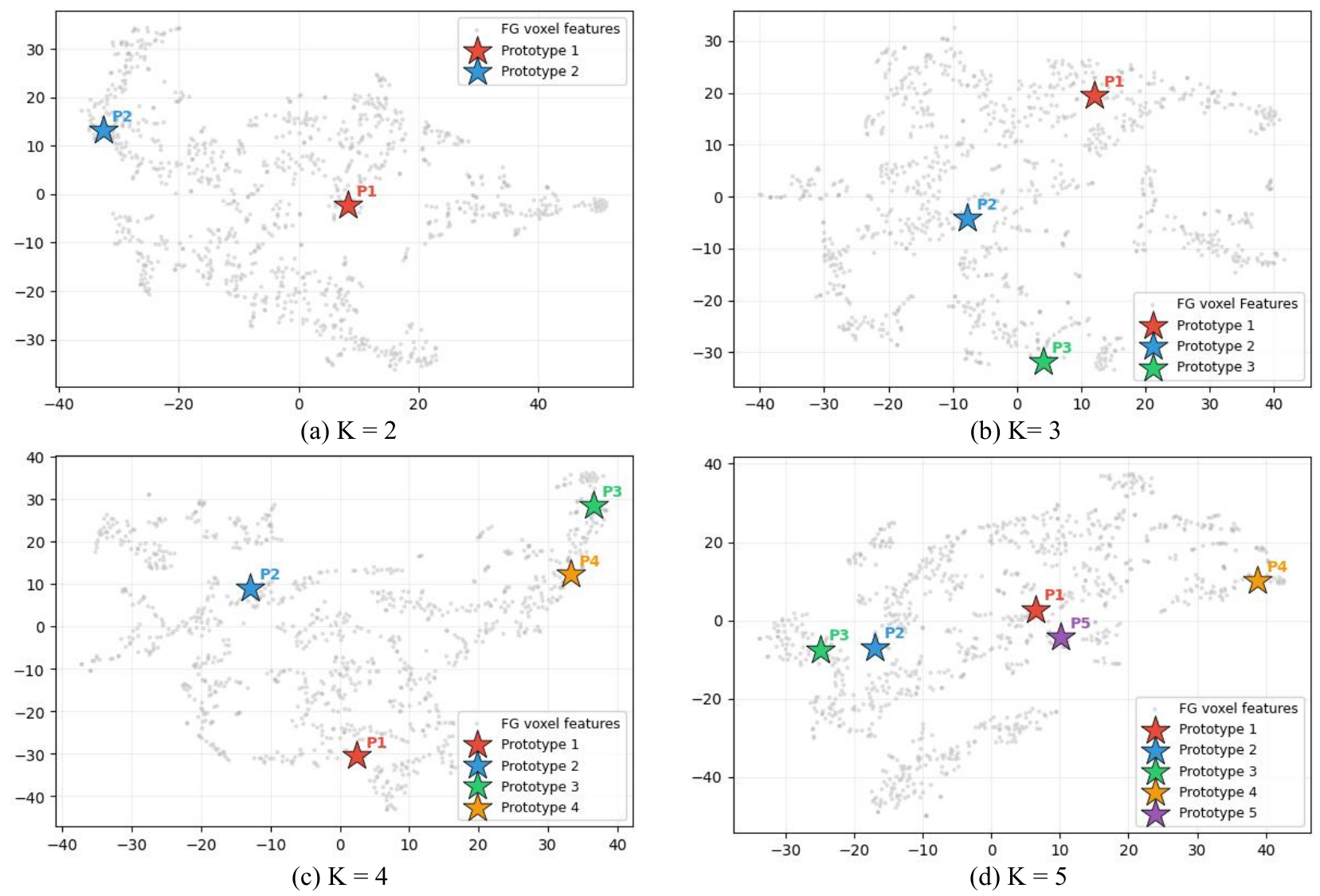}
    \caption{t-SNE visualization of \textbf{F}ore\textbf{G}round voxel features and corresponding prototype features under varying prototype numbers ($K=2$ to $K=5$) on the same BraTS2019 sample as Fig.~\ref{GMMnumber}, demonstrating that prototypes become less discriminative as $K$ exceeds 3.}
    \label{tsnenumber}
\end{figure}

\begin{figure*}
    \centering
    \includegraphics[width=0.8\linewidth]{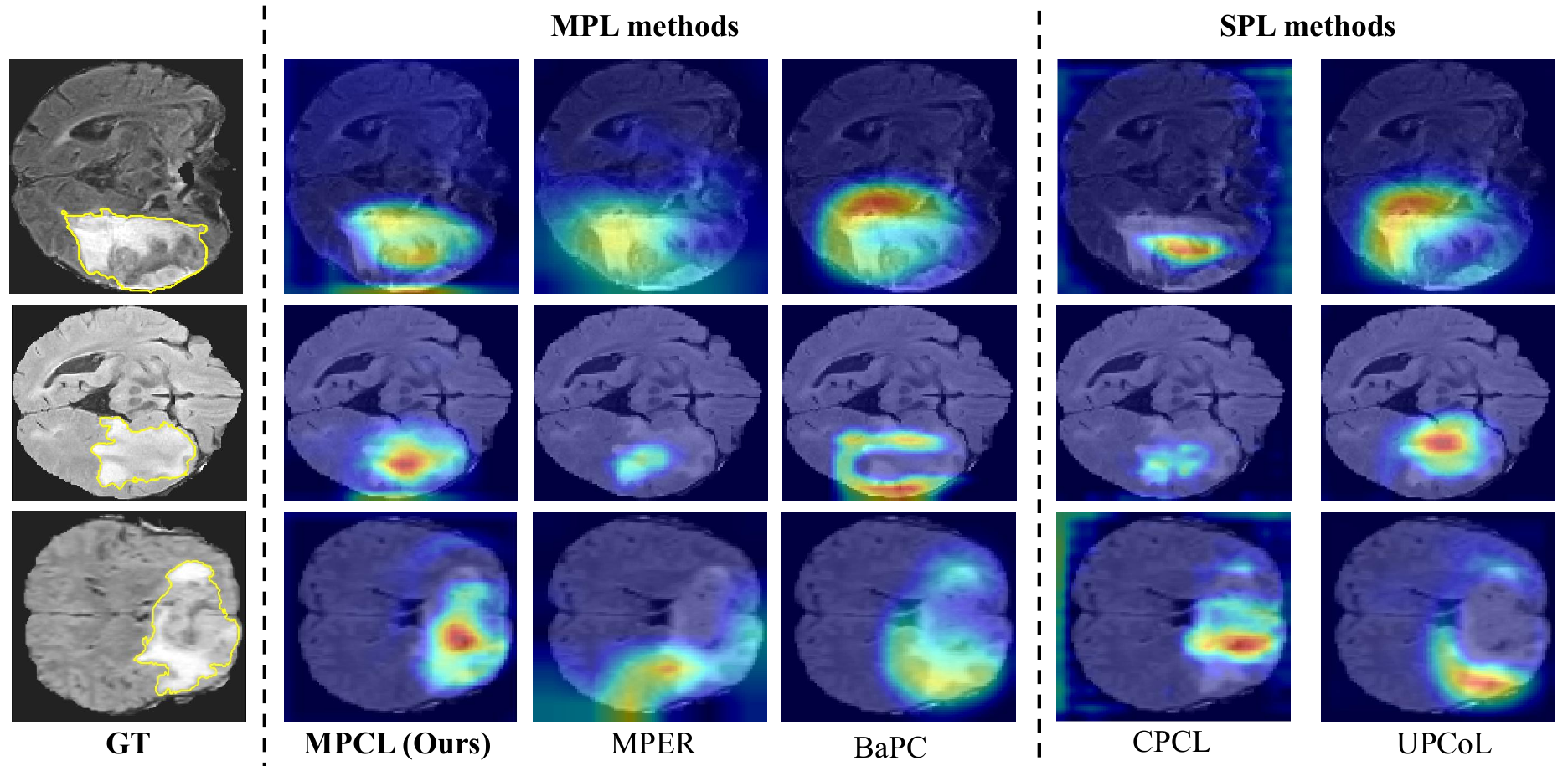}
    \caption{Grad-CAM visualization comparison between our MPCL and other PL SSMIS methods (including MPL methods MPER \cite{mper} and BaPC \cite{bapc}; SPL methods CPCL \cite{CPCL} and UPCoL \cite{upcol}) on BraTS2019 demonstrates that MPCL attends to more diverse intra-class heterogeneity structures while other MPL and SPL methods only capture partial structures. Ground truth is shown in yellow contours in the first column.}
    \label{camfigure}
\end{figure*}

To further investigate the impact of prototype number, we conduct a Pixel Intensity GMM decomposition analysis on BraTS2019 as shown in Fig.~\ref{GMMnumber}. As the component number in GMM decomposition increases beyond 3, the additional components begin to subdivide within existing intensity subregions, losing their representativeness of meaningful intensity-based subregions. This phenomenon is further reflected in the t-SNE visualization of prototype features generated via IHPG in Fig.~\ref{tsnenumber}: (a) provides insufficient prototype coverage of the foreground feature space; in (c) and (d), certain prototypes form redundant clusters in the embedding space, resulting in less discriminative representations. By contrast, (b) achieves the most balanced coverage across distinct feature regions. As a consequence, when $K > 3$, the additional prototypes selected by IHPG carry relatively less discriminative information, which marginally impacts the quality of prototype initialization and leads to a slight performance decline. Nevertheless, due to the subsequent optimization of PSO and DKA, the performance still remains competitive at $K > 3$.

\subsubsection{Analysis of Prototype Diversity}
To validate the effectiveness of our multiple prototype design, we provide Grad-CAM visualizations comparing MPCL with other PL methods (including MPL methods MPER \cite{mper} and BaPC \cite{bapc}; SPL methods CPCL \cite{CPCL} and UPCoL \cite{upcol}) in Fig.~\ref{camfigure}. The visualization reveals that MPCL attends to diverse structural regions across the entire anatomical structure, effectively capturing  intra-class heterogeneity structural representations. Among MPL methods, MPER shows attention patterns similar to SPL methods, concentrating on restricted regions, while BaPC attends predominantly to boundary areas.
Among SPL methods, CPCL focuses primarily on central regions while UPCoL attends mainly to boundary areas. All comparison methods fail to capture overall structural variations, whereas MPCL demonstrates comprehensive attention coverage across both central and boundary regions. This visualization demonstrates that MPCL effectively models the intensity-related intra-class heterogeneity inherent in medical images and provides more diverse representations.

\subsubsection{Analysis of Prototype Generation Strategies}
\begin{figure}
    \centering
    \includegraphics[width=1\linewidth]{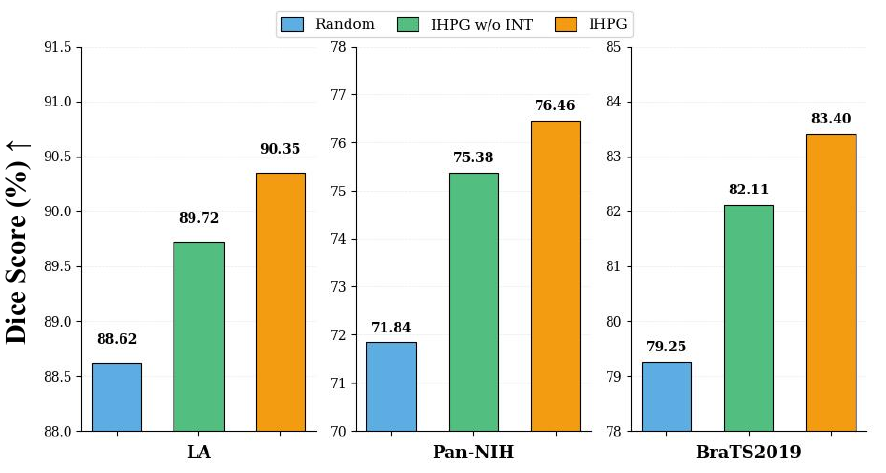}
    \caption{Dice score (\%) comparison of prototype generation strategies on three datasets under 10\% labeled data setting. IHPG with intensity-driven initialization (INT) consistently outperforms both random selection and random initialization (IHPG w/o INT), demonstrating the effectiveness of leveraging intensity characteristics for prototype generation.}
    \label{strategy}
\end{figure}

To validate the effectiveness of intensity-driven initialization in our IHPG, we compare three prototype generation strategies with the same number of prototypes (K=3) under 10\% labeled data setting, as shown in Fig.~\ref{strategy}. The three strategies are: Random selection, IHPG without INTensity-driven initialization (IHPG w/o INT), and our full IHPG method. The results demonstrate clear performance differences. 
Random selection achieves the lowest performance across all three datasets. This demonstrates that arbitrary prototype placement fails to capture meaningful structural patterns. IHPG w/o INT uses random initialization combined with our feature-spatial clustering strategy and achieves substantially better results. This validates that clustering spatially neighboring features is beneficial for capturing local structural characteristics, even without intensity guidance. Our full IHPG with intensity-driven initialization achieves the best performance with gains of +1.73\%, +4.62\%, and +4.15\% Dice over Random selection respectively. The consistent improvements demonstrate that both our intensity-driven initialization and feature-spatial clustering mechanism effectively generate more heterogeneous prototypes that provide representations with better diversity, achieving more precise segmentation results.

\subsubsection{Analysis of Sensitivity to $\lambda(s)$}
MPCL shows moderate sensitivity to the $\lambda(s)$ scheduling strategy, where our Gaussian warming-up function consistently achieves the best results. This demonstrates the effectiveness of our training-adaptive warming-up design. We compare three different schedules: (i) constant $\lambda(s)=1$, (ii) linear warming-up function $\lambda(s)=s/s_{\max}$, and (iii) Gaussian warming-up function $\lambda(s)=e^{-5(1-s/s_{\max})^{2}}$ (ours), as shown in Table~\ref{ablationlambda}. The consistently worst performance under (i) confirms that a fixed constraint weight fails to adapt to varying training dynamics, making progressive warming-up a robust and necessary design principle. Comparing (ii) and (iii), the latter achieves the best performance across all datasets, as it suppresses the influence of unlabeled predictions in early training to accommodate network instability, then rapidly increases as the network stabilizes, better aligning with the underlying training dynamics than the linear schedule.

\begin{table}[]
\caption{Ablation study on different $\lambda(s)$ schedules under 5\% labeled data setting, where (i) $\lambda(s)=1$, (ii) $\lambda(s)=s/s_{\max}$, and (iii) $\lambda(s)=e^{-5(1-s/s_{\max})^{2}}$ (ours), demonstrating the effectiveness of our training-adaptive warming-up design. The best results are highlighted in bold.}
\label{ablationlambda}
\centering
\renewcommand{\arraystretch}{1.2}
\resizebox{0.8\linewidth}{!}{
\begin{tabular}{c|c|cccc}
\bottomrule[1pt]
\textbf{Dataset} & \textbf{$\lambda(s)$} & \textbf{Dice↑} & \textbf{Jaccard↑} & \textbf{95HD↓} & \textbf{ASD↓} \\ \hline
\multirow{3}{*}{LA} & (i) & 84.90 & 74.41 & 10.73 & 3.19 \\
                    & (ii) & 85.60 & 75.23 & 9.77 & 2.79 \\
                    & (iii) & \textbf{86.91} & \textbf{77.15} & \textbf{7.97} & \textbf{2.48} \\ \hline
\multirow{3}{*}{Pan-NIH} & (i) & 54.35 & 39.98 & 35.92 & 5.80 \\
                         & (ii) & 57.64 & 41.93 & 27.10 & 3.94 \\
                         & (iii) & \textbf{59.01} & \textbf{43.72} & \textbf{23.85} & \textbf{3.16} \\ \hline
\multirow{3}{*}{BraTS2019} & (i) & 78.53 & 67.79 & 13.02 & 2.08 \\
                           & (ii) & 79.50 & 69.02 & 12.45 & 2.06 \\
                           & (iii) & \textbf{80.10} & \textbf{69.35} & \textbf{11.86} & \textbf{1.97} \\
\toprule[1pt]
\end{tabular}}
\end{table}

\section{Conclusion and Discussion}
In this work, we propose Multiple Prototype Contrastive Learning (MPCL), a novel multiple prototype learning framework in SSMIS, aiming to provide a novel paradigm for leveraging intra-class heterogeneity in medical images. By embracing intra-class heterogeneity of medical images, we obtain more diverse structural representations and achieve more precise segmentation results. It consists of three novel designs: (1) IHPG effectively models intra-class heterogeneity by generating multiple prototypes aligned with intensity characteristics, providing structural representations with better diversity. (2) PSO systematically optimizes a more discriminative and generalizable prototypical space, further enhancing more diverse structural representations and building a solid foundation for more precise segmentation. (3) DKA efficiently promotes intra-class heterogeneity knowledge transfer from prototypical space to the segmentation network, achieving segmentation results with better precision.
Moreover, extensive experiments on three medical image datasets with significant intra-class heterogeneity  demonstrate that MPCL significantly outperforms existing methods, especially under extremely limited labeled data. Our code is publicly available at https://github.com/rhodaliu17/MPCL.

\textbf{Discussion of Limitation}: 
One potential limitation of MPCL lies in modeling intra-class heterogeneity in low-dose imaging. In such cases, elevated noise levels obscure the underlying intensity distribution within target structures, which causes the intensity peaks identified by our IHPG to become less distinctive and correspondingly reduces the performance gain from intensity alignment. 
This limitation can be mitigated by integrating denoising as a preprocessing step prior to MPCL, or by extending MPCL with adaptive prototype number selection that adjusts to the degraded intensity distribution. 
Our future work will also be devoted to enhancing the robustness of 
intensity-based heterogeneity modeling under more challenging imaging conditions.

\textbf{Discussion of Future Work}: 
MPCL has demonstrated strong performance in leveraging intensity-manifested intra-class heterogeneity for SSMIS. Therefore, our future work will explore the generalization of intensity-aligned prototypes to broader medical image analysis tasks, such as classification. In the classification setting, intensity-aligned prototypes can provide fine-grained representations of heterogeneous subregions within each diagnostic category, offering more diverse and discriminative cues for accurate classification decisions.

\section*{Acknowledgement}
This work was supported by the National Natural Science Foundation of China (No. 62472315, No. 62476165).

\bibliographystyle{unsrt}

\bibliography{cas-refs}



\end{document}